\PassOptionsToPackage{prologue,dvipsnames, table}{xcolor}

\documentclass[10pt,twocolumn,letterpaper]{article}
\usepackage[accsupp]{axessibility} 
\usepackage{comment}
\usepackage{wacv}              

\definecolor{wacvblue}{rgb}{0.21,0.49,0.74}
\usepackage[pagebackref,breaklinks,colorlinks,allcolors=wacvblue]{hyperref}

\usepackage{adjustbox}

\usepackage{graphicx}
\usepackage{amsmath}
\usepackage{amssymb}
\usepackage{booktabs}
\usepackage{comment}

\usepackage{times}
\usepackage{epsfig}
\usepackage{float}
\usepackage{multirow}
\usepackage{amsfonts}
\usepackage{graphics}
\usepackage{witharrows}
\usepackage{arydshln}
\usepackage{amsfonts} 
\usepackage{mathtools}
\usepackage{enumitem}

\usepackage{algorithm}
\usepackage{algpseudocode}

\def\wacvPaperID{2205} 
\def\confName{WACV}
\def\confYear{2026}

\title{Pretraining Helps When Capacity Allows: \\ Evidence from Ultra‑Small ConvNets} 

\author{Srikanth Muralidharan \qquad Heitor R. Medeiros \qquad Masih Aminbeidokhti \\ Eric Granger \qquad Marco Pedersoli \\
LIVIA, Dept. of Systems Engineering, ETS Montreal, Canada\\
International Laboratory on Learning Systems (ILLS)}

\begin{document}
\maketitle
\begin{abstract}
Robust visual recognition on embedded platforms requires models that both generalize out‑of‑distribution (OOD) and fit into tiny compute/memory budgets. While pre‑training is a standard route to robustness for mid/large backbones, its value in the ultra‑small regime remains unclear. We present a capacity‑aware study of pre‑training for two efficient ConvNet families (EfficientNet and MobileNetV3) scaled from “small” to “ultra‑small” via a simple, reproducible recipe. We compare three initializations — ImageNet$\rightarrow$COCO pretraining, ImageNet classification pretraining, and training from scratch—on two axes of distribution shift: (i) cross‑dataset RGB$\rightarrow$RGB transfer between LLVIP and FLIR (ii) cross‑modality detection where models are fine‑tuned on RGB and evaluated on infrared (IR). A complementary classification study on DomainNet probes whether the trends extend beyond detection. Across settings, we find that pretraining’s benefit is conditional on both backbone capacity and shift difficulty. Task‑aligned Imagenet$\rightarrow$COCO  pretraining is the most reliable starting point at moderate sizes and for the easier transfer direction. In the low‑capacity regimes, differences are typically within run‑to‑run variation, and training from scratch can match or surpass pre‑training. Classification mirrors this capacity gating. Our results test the premise "pretraining always helps" and instead quantify when task‑aligned pretraining pays off for ultra‑small backbones and when it likely does not\footnote{\url{https://github.com/srikanth-sfu/wacv26-ultrasmall-cnns}}.

\end{abstract}

\section{Introduction}

Visual recognition supports applications from autonomous driving and robotics to security, medical imaging, and remote sensing \cite{krizhevsky2012imagenet,janai2020computer,litjens2017survey,cheng2017remote}. Two properties are crucial for deployment: generalization across environments, sensors, and conditions, and efficiency, enabling real-time inference on embedded devices. While generalization \cite{michaelis2019benchmarking} and efficiency \cite{tan2019efficientnet,koonce2021mobilenetv3} have been widely studied, their joint pursuit—generalizable yet ultra-small models remains underexplored. Generalization is often achieved through large-scale pretraining followed by fine-tuning \cite{girshick2014rich,girshick2015fast}, which reliably benefits high-capacity models \cite{kornblith2019better,yosinski2014transferable,he2019rethinking}. At the other extreme, classical convex settings suggest little dependence on initialization \cite{boyd2004convex}. The poorly understood regime lies in between: \textbf{how much does pretraining help when shrinking modern vision backbones to the ultra-small (sub-million-parameter) scale?}

\begin{figure}[t]
\centering
\includegraphics[width=\linewidth]{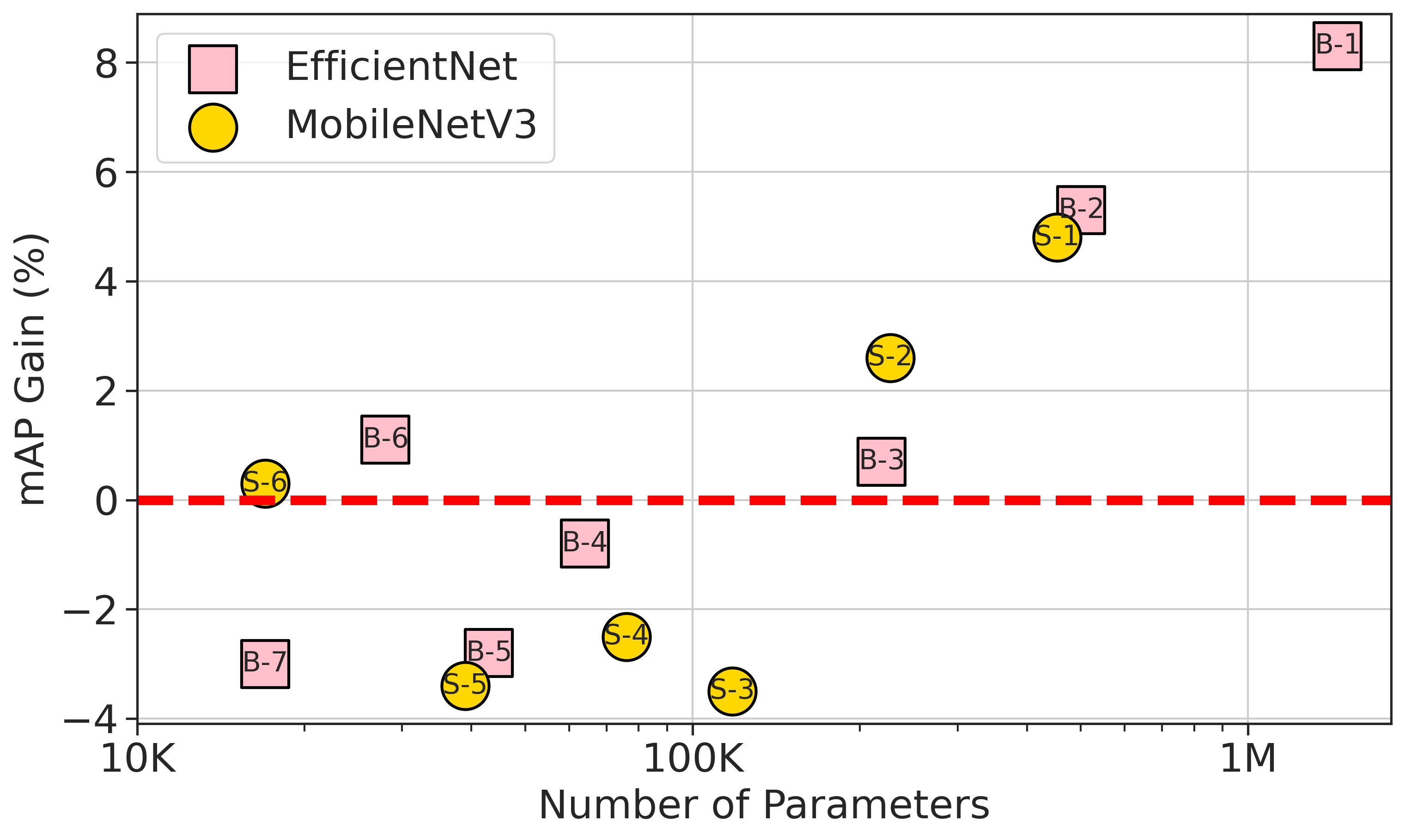}
\caption{\textbf{Out‑of‑distribution mAP gain from ImageNet pre‑training on FLIR for ultra‑small models.} The x‑axis reports model size (log‑scaled parameters); the y‑axis reports mAP when the detector is fine‑tuned on RGB and evaluated on IR. The red dashed line marks zero benefit: points above indicate a positive gain over random initialization. For backbones with $\gtrsim$100k parameters, the pre‑training advantage tends to grow, suggesting a monotonic link between capacity and cross‑domain generalization. In contrast, smaller networks hover unpredictably around zero, revealing no consistent trend at ultra‑low parameter counts.}
\label{fig:memory}
\end{figure}

We examine this question with two efficient ConvNet families (EfficientNet, MobileNetV3), three initializations (COCO, ImageNet, scratch), and two shift types: \textbf{cross-modality} (RGB$\!\rightarrow$IR) and \textbf{cross-dataset} (RGB$\!\rightarrow$RGB). Our main application is object detection, motivated by embedded night-time perception \cite{rogalski2017infrared}, complemented by domain generalization for classification to assess generality of our findings beyond our principal application. To densify the capacity axis, we extend EfficientNet (B7$\ldots$B0) with B$-1\ldots$B$-7$ and MobileNetV3 with S$-0\ldots$S$-6$ (Table~\ref{tab:sizes}).

Our study shows that pretraining benefits are conditional, not universal.  
For cross-dataset generalization for object detection ImageNet-COCO pretraining (also referred to as IN$\rightarrow$COCO pretraining) usually outperforms ImageNet pretraining (also referred to as IN) and scratch, but effects shrink with capacity and can vanish in harder directions (e.g., LLVIP$\!\rightarrow$FLIR). For cross-domain generalization IN$\rightarrow$COCO is most reliable, especially on FLIR, though scratch competes at the smallest sizes. For classification, IN improves large to moderate sized models, but gains fade or reverse for smallest of model sizes. Our additional controlled experiments on impact of fine-tuning data size confirm the same capacity-dependent patterns (Fig.~\ref{fig:memory}). Our findings differ from common notion that supervised pretraining recipes are necessary to obtain better generalization regardless of the model properties. Benefits depend jointly on model capacity and shift difficulty: clear gains appear for moderate sizes or easier shifts, but drop to noise below a certain threshold for difficult transfers.

\paragraph{Our contributions can be summarized as follows:}
\begin{itemize}
\item A capacity-aware evaluation of EfficientNet and MobileNetV3 across detection and classification, with IN$\rightarrow$COCO, IN, and scratch initializations.
\item New ultra-small variants (B$-1\ldots$B$-7$, S$-0\ldots$S$-6$) and an explicit scaling recipe for reproducibility.
\item Empirical evidence that pretraining benefits diminish with extreme capacity reduction, sometimes making random initialization competitive.
\item Practical guidance for embedded deployment: when pretraining is worthwhile versus when task-specific data tuning matter more.
\end{itemize}

Our aim is not to propose a new backbone but to quantify capacity - pretraining interactions under embedded constraints, providing actionable evidence for design at the low-capacity frontier.

\begin{table}
    \centering
    \caption{The ultra-small microcontrollers we target deliver barely one-sixth of the compute available on a modest Raspberry Pi-class processor, and their tiny on-chip memory leaves almost no headroom for model weights or intermediate activations. In this work we extended the common EfficienNet Models (from B7 to B0) with 7 additional tiny and ultra-small models (B-1 to B-7) to be used in our experiments.}
    \resizebox{\columnwidth}{!}{
    \begin{tabular}{ccccc}
    \toprule
    \multirow{2}{*} & \textbf{Commercial GPU} & \textbf{Mobile} & \textbf{Tiny} & \textbf{Ultra-Small} \\
    & (NVIDIA RTX)   & (iPhone 16)  & (Cortex-A53) & (Cortex-M7)   \\
    \toprule
        \textbf{Models} & B7 to B4 & B4 to B1 & B0 to B-2 & B-3 to B-7 \\
        \textbf{FLOPs} & $\sim$82T & $\sim$2.15T & $\sim$12.3G & $\sim$2.0G \\ 
        \textbf{Memory} & 24GB & 8GB & 2GB & 1GB \\
    \bottomrule
    \end{tabular}
    }
    \label{tab:sizes}
\end{table}
\section{Related Work}

\noindent \textbf{Efficient ConvNets.} Different methods were proposed to focus on efficiency and low-cost CNNs. Methods like MobileNetV1~\citep{howard2017mobilenets} leveraged depth-wise separable convolutions with adjustable complexity through width and resolution multipliers, while MobileNetV2~\citep{sandler2018mobilenetv2} introduced inverted residual layers and linear bottlenecks to further preserve information flow and reduce computation. More recently, models such as MobileNetV3~\citep{howard2019searching} and MnasNet~\citep{tan2019mnasnet} have employed Neural Architecture Search (NAS), explicitly incorporating latency as an optimization criterion. Alternative approaches, including ShuffleNet~\citep{zhang2018shufflenet} and GhostNet~\citep{han2020ghostnet}, introduced novel convolutional operations, such as channel shuffling, group convolutions, and Ghost modules, to significantly reduce parameters and computational cost. Additionally, transformer-based methods such as MobileViT~\citep{mehtamobilevit} and TinyViT~\citep{wu2022tinyvit} have effectively integrated global context processing into compact architectures, demonstrating superior accuracy and transferability across multiple vision tasks. 
Even if those model families are efficient and thought for embedded devices, they still contain millions of parameters and cannot be used directly on micro-controllers that are commonly used in industry (see Tab.~\ref{tab:sizes}). For this reason, for our experiments, we extend two families of models (EfficientNet and MobileNetV3) to much smaller sizes down to the order of 10k parameters.
\\

\textbf{Out-of-distribution Robustness.}
Robustness to distribution shifts remains a critical challenge in object detection, particularly in safety-critical applications. Prior works have studied domain adaptation~\cite{saito2019strong}, data augmentation~\cite{ghiasi2021simple,yun2019cutmix}, and self-training~\cite{zoph2020rethinking} to improve generalization across domains. Recent efforts focus on detecting out-of-distribution images~\cite{yang2024generalized} and improving model calibration under distribution shift~\cite{ovadia2019can}. \\
Pretraining is a common practice in tasks such as image classification and object detection, significantly boosting performance, especially for larger models~\cite{girshick2014rich,redmon2018yolov3}. It has been shown to accelerate convergence and enhance accuracy when transferring to tasks with limited training data~\cite{girshick2014rich,kornblith2019better}. Traditionally, ImageNet pretraining has been the standard, providing robust features that transfer well to various downstream tasks. Recent studies have explored self-supervised pretraining~\cite{caron2021emerging,he2022masked} to further improve downstream performance. However, these works predominantly focus on large or standard-scale models. The impact of pretraining on ultra-small models, particularly those with backbone size up to two orders of magnitude smaller than EfficientNet-B0~\cite{tan2019efficientnet}, remains underexplored. Our work investigates the OOD robustness of models that are orders of magnitude smaller than conventional architectures, addressing the need for efficient yet robust frameworks in resource-constrained deployment scenarios. \\

\section{Method}


\subsection{Preliminaries}


\paragraph{Model family generation.}
We systematically construct ultra-small models by downscaling two backbone networks, EfficientNet-B0~\cite{tan2019efficientnet} and MobileNetV3-small~\cite{howard2019searching}. Specifically, we reduce both the width and depth of these models up to seven levels in case of EfficientNet down EfficientNet-B0 and six levels down MobileNet-V3 small while maintaining input resolution, which is critical for downstream tasks such as object detection. Details of our downscaling procedure are provided in the supplementary material.


\subsection{Proposed generalization evaluation pipeline}
\begin{figure}[!h]
  \centering
  \includegraphics[width=1\linewidth]{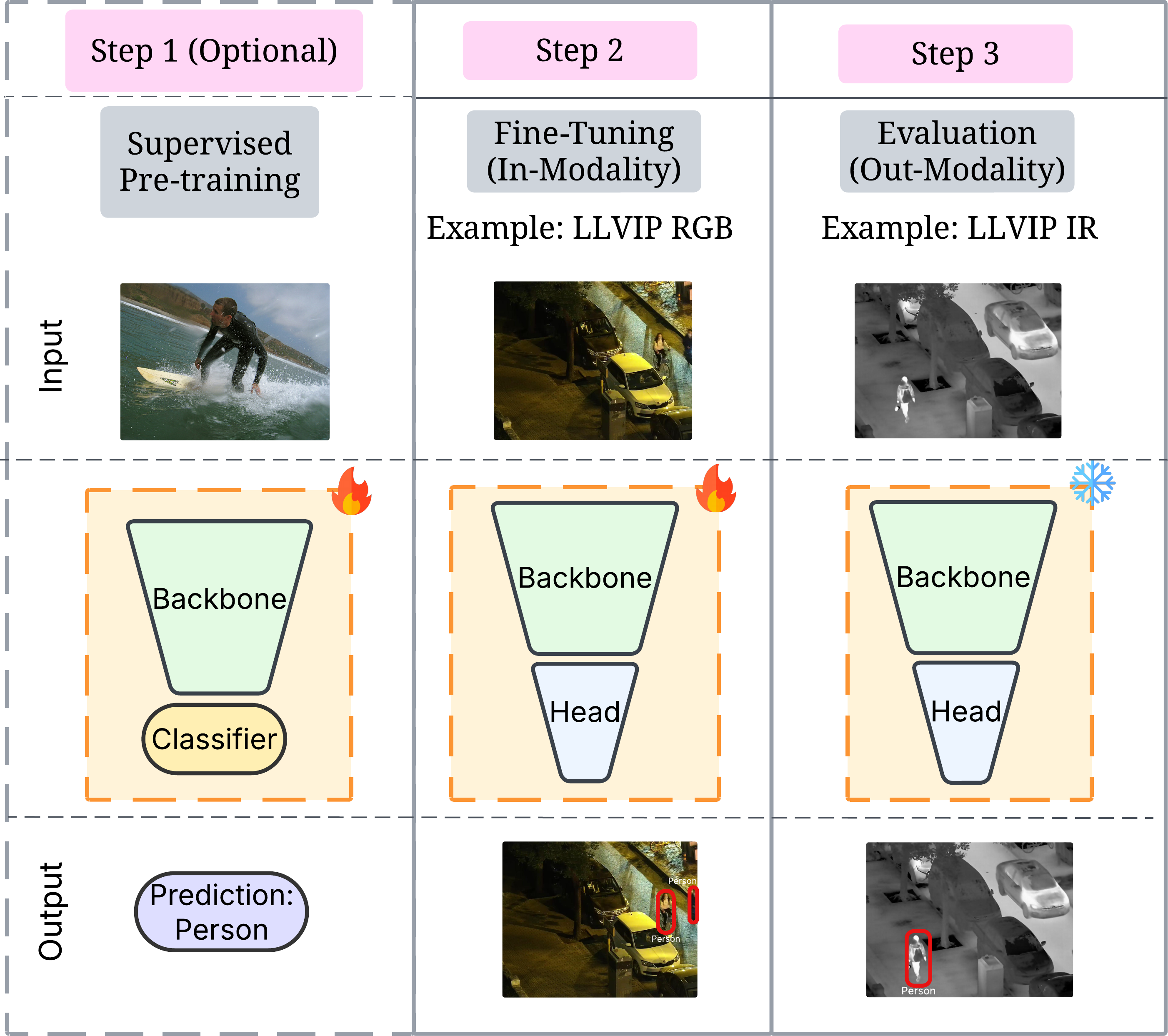}
  \caption{\textbf{Phases of our approach.} We obtain initialization weights for our model families in two ways: using supervised pretraining (e.g. Imagenet classification / COCO detection) and random initialization. We then train these models for classification or detection tasks with each initialization in parallel on an In-domain detection dataset. Finally, we evaluate effectiveness of supervised pretraining by testing both models on cross-modal and cross-domain object detection tasks.}
  \label{fig:figure2}
\end{figure}

Figure~\ref{fig:figure2} presents an overview of our framework. The pipeline comprises three main steps: (1) initializing the detector backbone either with ImageNet pretraining or with an additional optional task specific supervised finetuning such as with COCO dataset for object detection or random initialization, (2) finetuning models on in-domain data, and (3) evaluating out-of-domain generalization to compare the effectiveness of different pretraining against random initialization.

\paragraph{Supervised ImageNet pretraining (IN)}
\label{sec:imgnetpretrain}
Supervised pretraining on ImageNet~\cite{deng2009imagenet} has been a standard approach to initialize convolutional neural networks for various downstream tasks. In this setting, models are trained on the ImageNet-1K dataset, which contains approximately 1.28 million training images across 1000 object categories with standard recipe of using cross entropy loss. The pretraining process results in models with weights that capture generic visual representations, which are transferable to downstream tasks and domains. We adopt this standard supervised pretraining to initialize our backbones and compare them against no pretraining as well as task-specific pretraining (when applicable) for mutliple tasks including object detection and Image classification for various scales of ultra-small models. We refer to this pretraining paradigm as IN in subsequent sections, figures, and captions.

\paragraph{Supervised COCO finetuning(IN$\rightarrow$COCO)}
\label{sec:imgnetpretrain}
For object detection task, we further also consider supervised fine-tuning on the COCO detection dataset, after standard ImageNet-1K supervised backbone initialization~\cite{lin2014microsoft}. In this setting, models are trained on the COCO dataset with the NanoDet standard recipe~\cite{lyu2021nanodet}. The pretraining process results in models with weights that capture detection-specific representations, which are transferable to downstream object detection tasks that generalize across different domains and modalities. In this work, we assess this pretraining benefit for different levels of small models against task-agnostic pretraining and no pretraining variants. We refer to this two-stage detection-aligned pretraining paradigm as IN$\rightarrow$COCO in all subsequent sections, figures, and captions.

\paragraph{Finetuning Object Detectors.}
We study supervised object detection using data from an in-domain distribution $\mathcal{D}{in} = {(x_{in}, Y_{in})}$, where $x_{in} \in \mathbb{R}^{W_{in} \times H_{in} \times C}$ denotes an image with spatial dimensions $W_{in} \times H_{in}$ and $C$ channels. Our goal is to evaluate the generalization capability of object detectors when tested on multiple generalization tasks $\mathcal{D}_{out}$, comparing models initialized with random weights versus Imagenet pretrained weights, and examining how model size affects different types of generalization.
We train three versions of each detector:

\begin{itemize}
\item \textbf{Random initialization:} All layers are initialized from scratch without any pretraining.
\item \textbf{Task Independent large scale pretraining:} Backbone networks are initialized with IN pretrained weights.
\item \textbf{Task-specific pretraining:} Backbones are initialized with  task-specific pretraining paradigm, such as IN$\rightarrow$COCO for object detection, and are used to initialize most of the detector layers.
\end{itemize}

All the variants are trained using supervised in-domain data. This ensures that the detectors learn representations useful for both in-domain and generalization settings.

\paragraph{Generalization Evaluation.}
We assess the relationship between model size and pretraining generalization from multiple perspectives. First, we assess generalization cross-domain(s) for both object detection and image classification. Second, we test cross-modal generalization where a detector is fine-tuned on RGB images and then evaluated on infrared data. The third evaluation protocol involves viewpoint generalization (using our In-house Distech dataset) with multiple cameras/views. In this case, the fine-tuning is performed on some viewpoints, while the evaluations are run on viewpoints not seen during fine-tuning.
\section{Experiments}


\begin{table}[t]
\centering
\small
\setlength{\tabcolsep}{8pt}
\renewcommand{\arraystretch}{1.1}
\begin{tabular}{lccccc}
\toprule
\multirow{2}{*}{\textbf{Model}} & \multicolumn{2}{c}{\textbf{Backbone}} & \multicolumn{2}{c}{\textbf{Detector}} \\
\cmidrule(lr){2-3} \cmidrule(lr){4-5}
 & \textbf{Params} & \textbf{MFLOPs} & \textbf{Params} & \textbf{MFLOPs} \\
\midrule
\rowcolor[gray]{0.94}\multicolumn{5}{c}{\textbf{MobileNetV3 (input $384\times480$)}}\\
\midrule
S   & 927K  & 226.38 & 1.27M & 675.75 \\
S-1 & 453K  & 146.98 & 723K  & 500.41 \\
S-2 & 227K  & 109.99 & 470K  & 428.86 \\
S-3 & 118K  & 71.32  & 325K  & 344.01 \\
S-4 & 76K   & 56.78  & 271K  & 313.71 \\
S-5 & 39K   & 53.26  & 232K  & 307.01 \\
S-6 & 17K   & 46.15  & 184K  & 267.51 \\
\midrule
\rowcolor[gray]{0.94}\multicolumn{5}{c}{\textbf{EfficientNet (input $384\times480$)}}\\
\midrule
B0  & 3.6M  & 1459.77 & 3.87M & 1819.19 \\
B-1 & 1.45M & 798.19  & 1.67M & 1089.55 \\
B-2 & 501K  & 286.98  & 692K  & 539.40 \\
B-3 & 219K  & 152.01  & 391K  & 378.16 \\
B-4 & 64K   & 94.44   & 223K  & 304.09 \\
B-5 & 43K   & 89.29   & 193K  & 289.00 \\
B-6 & 28K   & 84.75   & 174K  & 278.09 \\
B-7 & 17K   & 81.97   & 160K  & 271.99 \\
\bottomrule
\end{tabular}
\caption{\textbf{Complexity of detector backbones and full detectors at input $384\times480$.}
Blocks follow the template style with grouped headers: the first block lists \emph{MobileNetV3} variants (S$\rightarrow$S-6) and the second lists \emph{EfficientNet} variants (B0$\rightarrow$B-7). \textbf{Params} are in K/M and \textbf{MFLOPs} report multiply–accumulate counts.}
\label{tab:model_params}
\end{table}

\subsection{Datasets} 
For Image Classification in domain generalization settings, we use DomainNet~\cite{peng2019moment} benchmark. The benchmark includes diverse datasets and domains such as clipart, quickdraw, sketch, painting, infograph, and real images. Following standard practice, we designate one domain as the training environment and treat the remaining domains as test environments. This setup allows us to quantify the effect of pretraining on classification generalization under a vanilla ERM objective. Further, in our experimental setup, we use default configuration for ERM training from Cha et al.~\cite{cha2021swad} only replacing backbone with our candidate backbones. We report classification accuracies on held-out domains to evaluate the impact of pretraining versus training from scratch in the small-model regime.

For Object Detection, we explored two public RGB/IR benchmarking datasets: LLVIP and FLIR, as well as our inhouse Distech dataset which is fully in IR domain. \textbf{LLVIP:} The LLVIP dataset is a surveillance dataset composed of $12,025$ RGB/IR pairs of images of size $1280 \times 1024$ for training and $3,463$ pairs for testing, consisting of person class annotations. \textbf{FLIR ALIGNED:} For the FLIR dataset, we follow the settings provided by Zhang et al.~\cite{zhang2020multispectral}, which consists of $4,129$ aligned pairs in the training set and $1,013$ pairs in the test set. The FLIR images have resolution of $640\times480$. It contains three objects class annotations: bicycles, cars, and people. Notably, we remove the rare category of ``dog" objects following previous approaches~\cite{10209020}.
We perform cross-domain and cross-modal generalization experiments with these datasets. \textbf{DISTECH:} We study viewpoint generalization with our in-house dataset known as Distech dataset. This dataset consists of images obtained from 30 overhead infrared domain cameras from multiple rooms. We split the data into trainval and test where the latter data is fully separate from trainval images in terms of rooms, the domains in our case. We use images from six cameras that is setup in a room entirely different from the rest of the cameras for test split and the images from rest the 24 cameras for the trainval split. We have in total 4177 images for training, 1045 images for validation and 1467 images in our test set.

\subsection{Training details}
\noindent \textbf{ImageNet pretraining.}
We pretrain all EfficientNet and MobileNetV3 backbones on ImageNet using the FFCV framework~\cite{leclerc2022ffcv}, for 32 epochs at $384{\times}384$ resolution with SGD and standard data augmentation (random resized crop + horizontal flip). The full recipe (LR schedule, batch size, weight decay, momentum and other details) is provided in the supplementary material.

\medskip\noindent \textbf{COCO detection finetuning.}
For detection finetuning, we adopt a single-stage FCOS-style NanoDet~\cite{lyu2021nanodet} with a PANet/FPN neck (96 channels, 3 scales; strides $s{=}\{8,16,32\}$) and a two-layer NanoDetHead (ReLU6 + BN, shared cls/reg features), using \textit{reg\_max}$\,{=}\,7$ and an octave base scale of 5. Losses we apply are Quality Focal Loss for classification, Distribution Focal Loss for box distribution regression, and GIoU for localization. We train at $480{\times}384$ resolution. Optimizer/schedule, batch size, and augmentation (scaling, translation, color, etc.) hyperparameters are detailed in the supplementary material.

\noindent \textbf{Object detector training on target domain RGB data:}  
We adopt the single stage FCOS sytle object detector from Nanodet~\cite{lyu2021nanodet} for all our experiments. For LLVIP and FLIR datasets, we retain FPN same as detection pretraining model configuration. For our in-house distech dataset, we skip FPN and feed the feature map with stride 8 directly to the head. The loss function is also same as that of COCO finetuning loss functions. \\
\noindent For data augmentation, we apply horizontal flip with a probability of $0.5$, translation with a probability of $0.2$, random scaling between $0.9$ and $1.1$. In case of LLVIP and FLIR datasets we train our models with batch size of $24$, use AdamW~\cite{loshchilov2017decoupled} optimizer with linear warmup for 800 steps and CosineAnnealing~\cite{loshchilov2016sgdr} learning rate scheduler. We train Object detector with the LLVIP dataset for $32$ epochs and in FLIR for $80$ epochs. For our inhouse distech dataset, we train for 120 epochs use Adam with warmup for 500 steps and StepLR schedule. For all datasets we use base learning rate of 0.005 except for the backbone in the pretrained variants for which we use base learning rate of 0.0005. \\

\noindent \textbf{(d) Baseline Methods:}
We use eight models from EfficientNet family and seven models from MobileNet-V3 family for our experiments. The number of MFLOPS and params of the family of all our models that are part of the detector are shown in tab.~\ref{tab:model_params}. As shown in the table our backbones in both the families have number of parameters starting from 17k and the biggest model we have is EfficientNet-B0 which has 3.6M parameters.

\subsection{Results}
In this section we present our quantitative results. We present results on LLVIP and FLIR about the ability of a model fine-tuned on RGB to detect objects on cross-dataset and cross-domain generalization (infrared data) from the same dataset. We further evaluate on the Distech dataset, the capabily of a model fine-tuned on infrared data, to adapt to new cameras. We also present findings beyond the Object Detection setting to domain generalization for image classification. Finally, we re-visit object detection and perform ablations on the amount of training data used for fine-tuning the model and the pretraining and fine-tuning resolutions used. We refer the reader to the supplementary material where we also report detailed results.

\begin{figure}
    \centering
    \hspace{-.5cm}
    \includegraphics[width=0.5\textwidth]{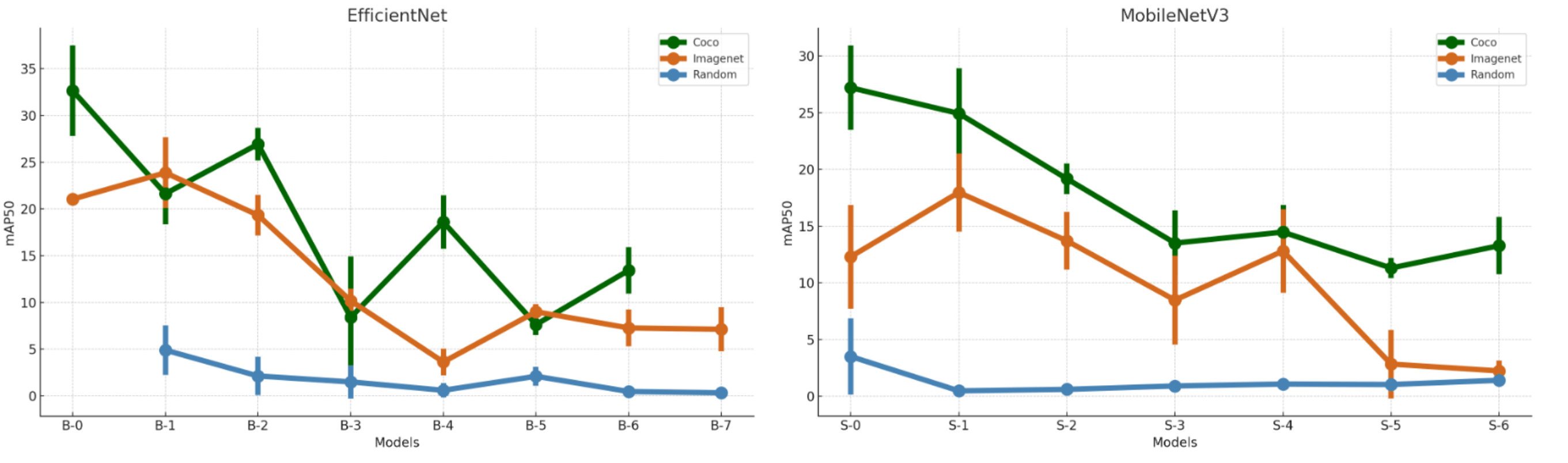}
    \caption{\textbf{Cross-dataset generalization from FLIR to LLVIP RGB images.} 
    Results are reported for two model families, \emph{EfficientNet} (left) and \emph{MobileNetV3} (right), across different model variants ($b{=}0$–$7$ and $s{=}0$–$6$). Curves compare three initialization strategies: \emph{IN$\rightarrow$COCO}, \emph{IN}, and \emph{Random}. Performance trends show that IN$\rightarrow$COCO pretraining paradigm consistently yields stronger generalization, while Random initialization performs worst, especially for smaller model variants.}
    
    \label{fig:FLIR_LLVIP}
\end{figure}

\begin{figure}
    \centering
    \hspace{-.5cm}
    \includegraphics[width=0.5\textwidth]{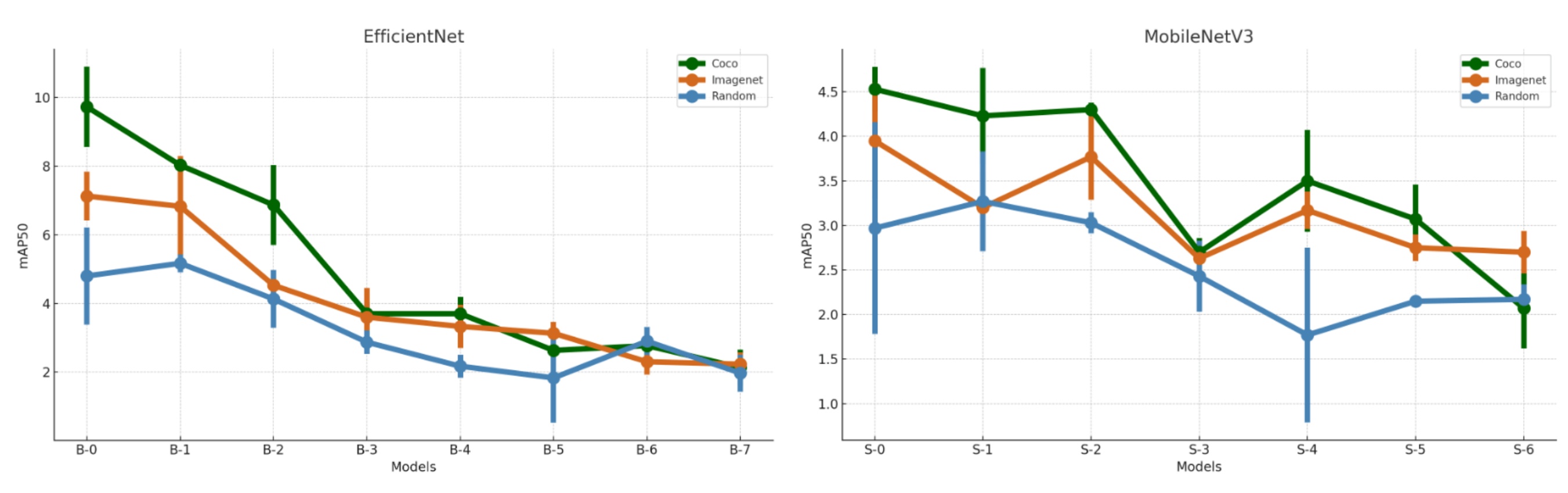}
    \caption{\textbf{Cross-dataset generalization from LLVIP to FLIR RGB images.} 
    Performance is shown for two model families, \emph{EfficientNet} (left) and \emph{MobileNetV3} (right), across different model variants ($b{=}0$–$7$ and $s{=}0$–$6$). Curves compare three initialization strategies: \emph{IN$\rightarrow$COCO}, \emph{IN}, and \emph{Random}. Results indicate that larger model variants tend to generalize better, with IN$\rightarrow$COCO pretraining paradigm providing the most consistent improvements.}
    
    \label{fig:LLVIP_FLIR}
\end{figure}

\paragraph{Cross-dataset generalization (RGB$\rightarrow$RGB) and the role of capacity.}
We evaluate transfer between LLVIP and FLIR using EfficientNet and MobileNetV3 backbones (Fig.~\ref{fig:FLIR_LLVIP}, \ref{fig:LLVIP_FLIR}). Across both families we observe two consistent phenomena. First, the benefit of IN$\rightarrow$COCO over IN or training from scratch depends strongly on capacity: for larger backbones (B-0..B-2, S-0..S-2) IN$\rightarrow$COCO finetuning confers clear improvements, whereas for the smaller half of the capacity range (B-4..B-7; S-3..S-6) the average IN$\rightarrow$COCO - IN gap shrinks and occasionally flips sign (e.g., B-5, B-7, S-6). Second, transfer is highly directional: FLIR$\rightarrow$LLVIP yields much higher absolute \(\text{mAP}_{50}\) than LLVIP$\rightarrow$FLIR. For MobileNet, COCO pretraining consistently dominates ImageNet while for EfficientNet the average gain remains positive but is less uniform. Taken together, these results indicate a capacity threshold below which pretraining alone does not reliably overcome dataset shift, and above which IN$\rightarrow$COCO pretraining paradigm is better.

Our cross-dataset generalization study involves RGB$\rightarrow$RGB  transfer between LLVIP and FLIR. Within this scope, the data supports the following prescriptions: (i) employ detection-style pretraining when using moderate-or-larger backbones (B-0..B-3 / S-0..S-2), where gains are consistent and exceed run-to-run variability; (ii) for small backbones (B-4..B-7 / S-3..S-6), pretraining provides at most marginal benefit on the harder LLVIP$\rightarrow$FLIR direction, suggesting that model capacity being more critical than the choice of pretraining. (iii) direction matters, with FLIR$\rightarrow$LLVIP consistently easier than the reverse. This shows that observed effects are properties of the domain shift rather than distinct architecture related or training-related features.

\begin{figure}[ht]
    \centering
    \begin{subfigure}[t]{0.5\textwidth}
        \centering
        \includegraphics[width=0.49\textwidth]{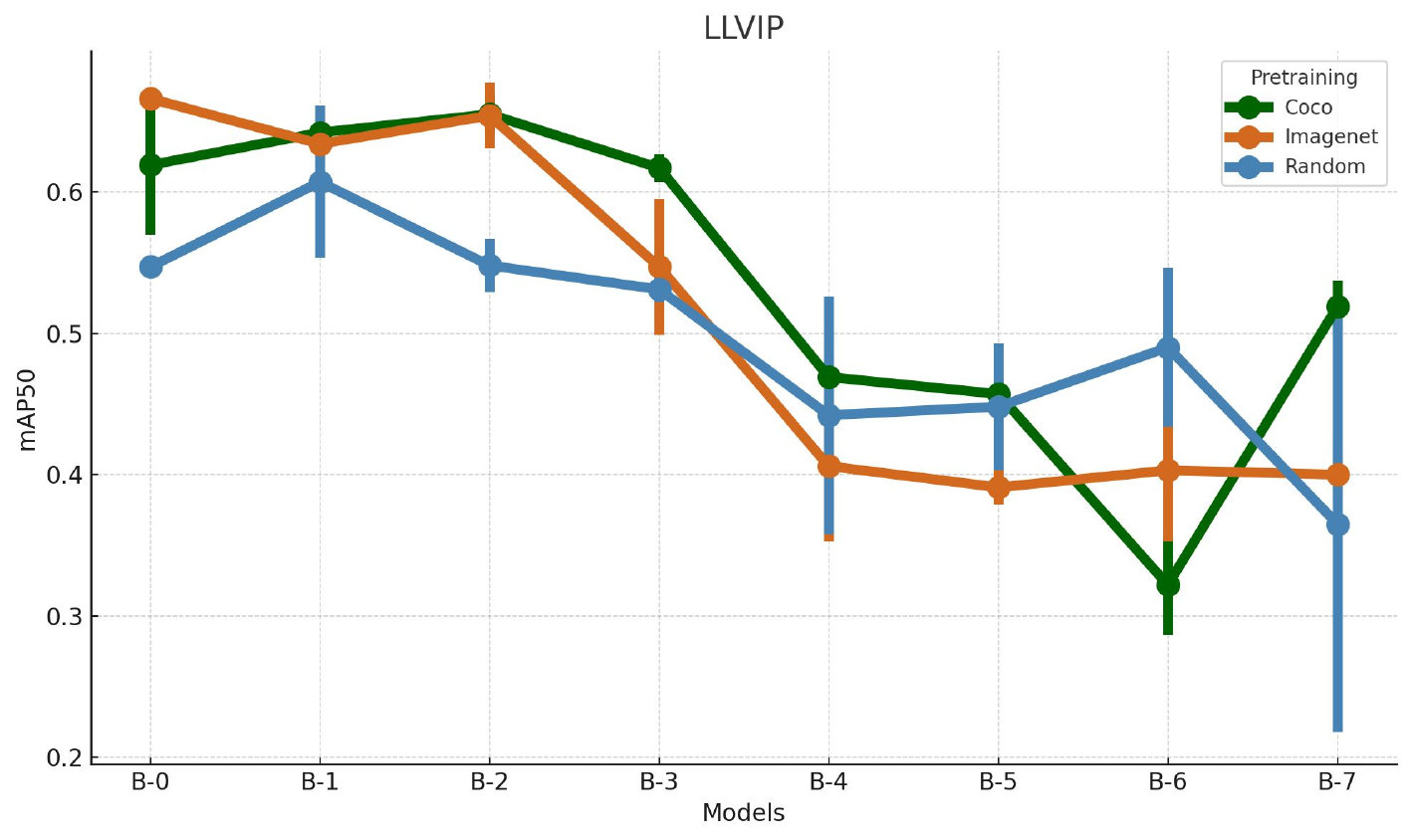}
    \hfill
        \includegraphics[width=0.49\textwidth]{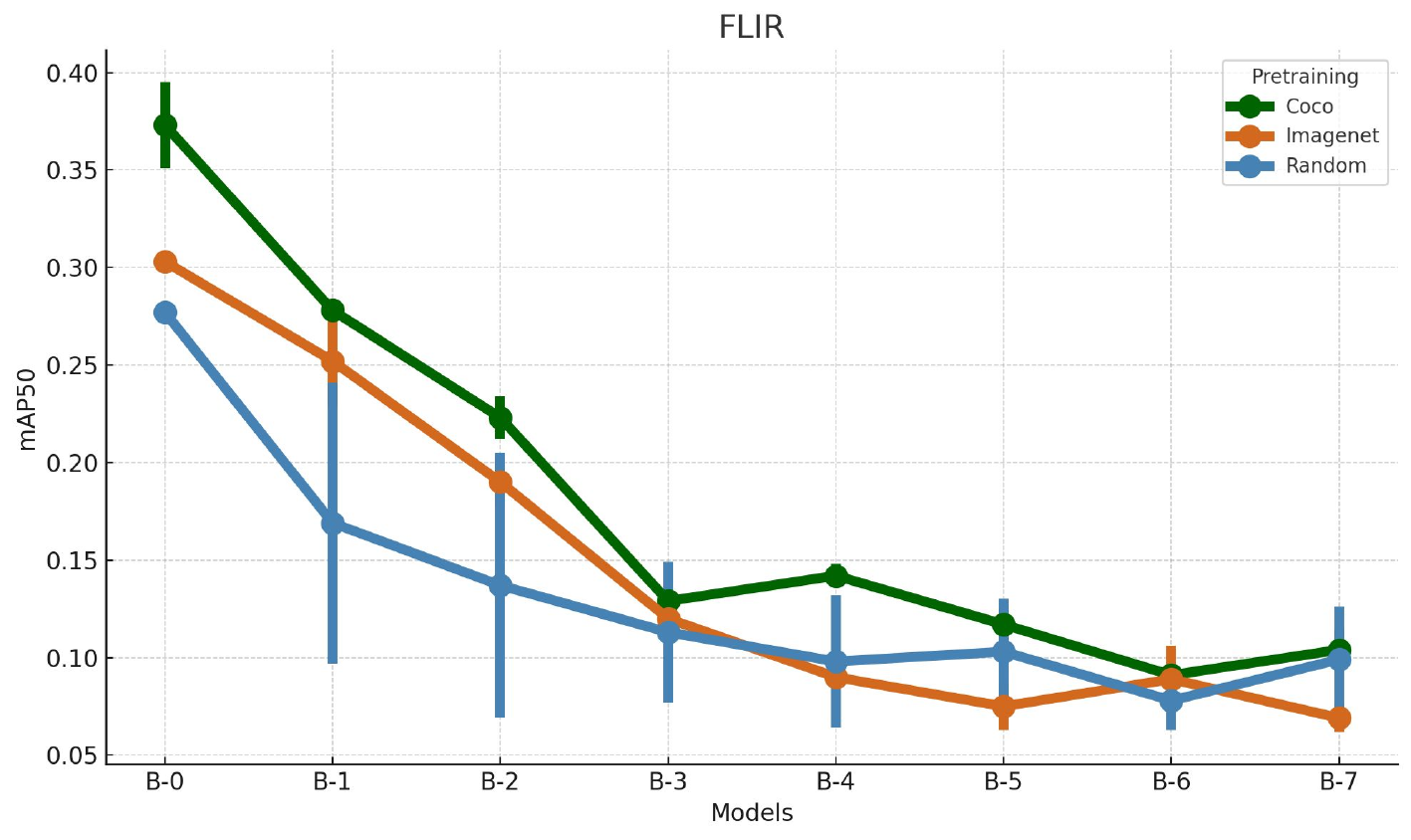}
        \caption{OOD EfficientNet.}
    \end{subfigure}
    \begin{subfigure}[t]{0.5\textwidth}
        \centering
        \includegraphics[width=0.49\textwidth]{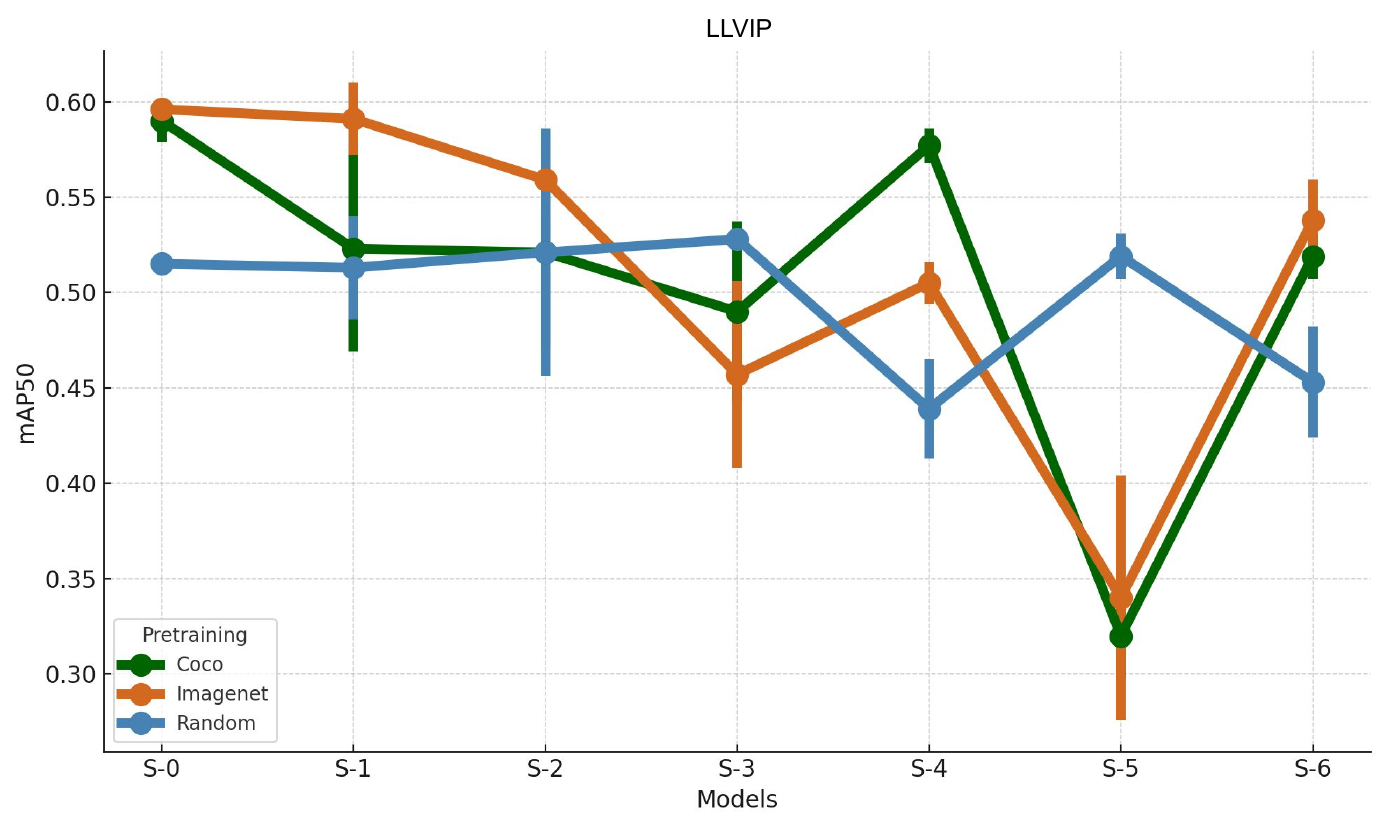}
    \hfill
        \includegraphics[width=0.49\textwidth]{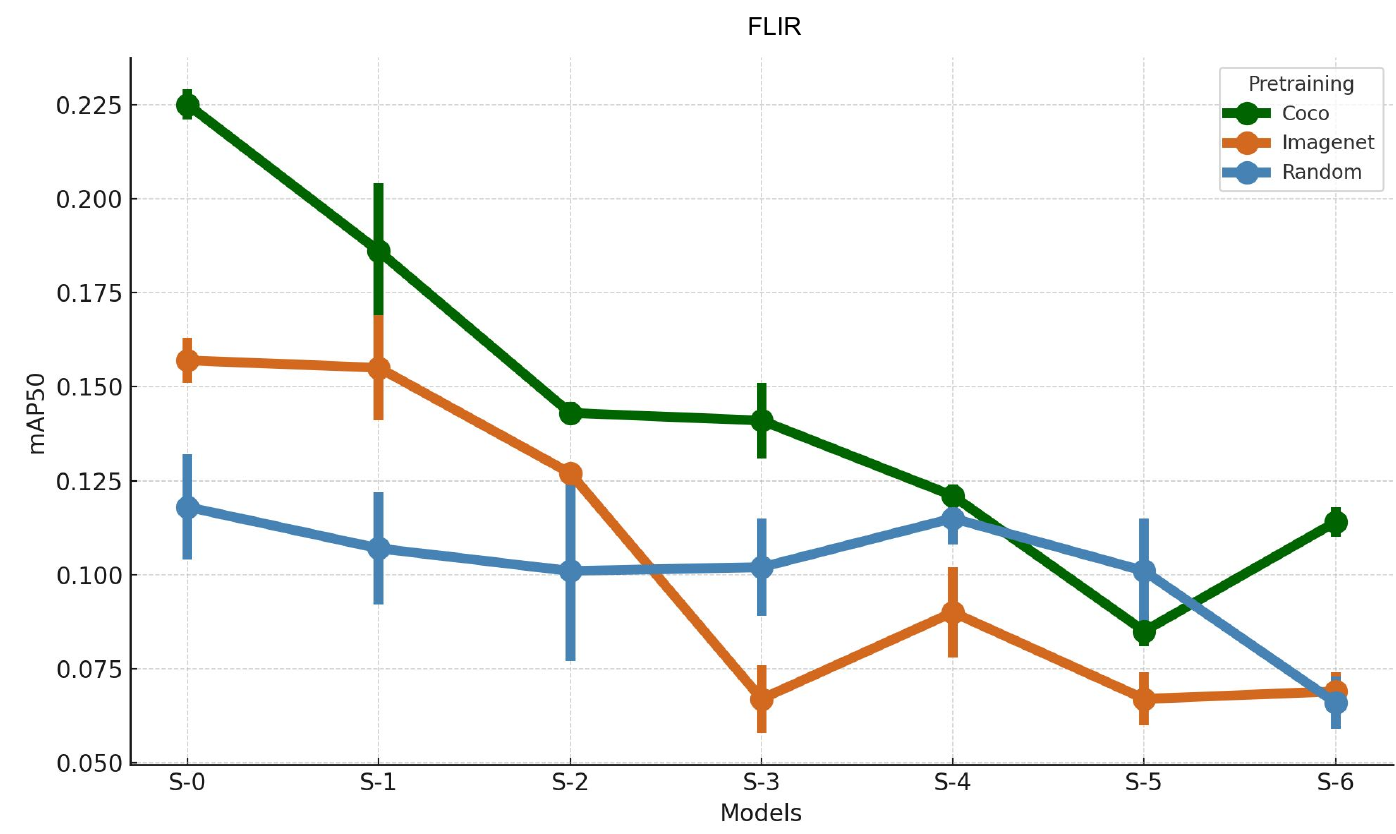}
        \caption{OOD MobileNetV3.}
    \end{subfigure}


    \caption{\textbf{Modality adaptation of ultra-small EfficientNet and MobileNet models from RGB to Infrared domain on LLVIP and FLIR datasets.} We observed that ImageNet pretraining is helpful only for the first few models for both families and datasets.}
    \label{fig:mainOOD}
\end{figure}

\paragraph{Modality Adaptation (RGB $\rightarrow$ Infrared):}
In Fig.~\ref{fig:mainOOD} we present generalization results on FLIR and LLVIP datasets with several ultra-small EfficientNet and MobileNet models. Here the task is to perform infrared detection with a model fine-tuned on RGB images of the same dataset. In this way, we make sure that the gap is only due to the different modality (RGB-Infrared) and no other changes of distributions are involved. Among the initialization strategies, IN$\rightarrow$COCO detection pretraining consistently provides the strongest starting point, even when the downstream task involves out-of-distribution infrared detection, especially larger models (B-0 to B-5 and S to S-2). 

Large models obtain a clear advantage from supervised pretraining on both datasets. This follows our intuition that pretraining helps until a certain model size. When the model becomes very small, pretraining does not help as the limited number of parameters cannot learn useful features from pretraining. To summarize, pretraining, specifically the task-aligned pretraining, remains useful for generalization for larger models, while its utility decreases with model size. 

\begin{figure}
    \centering
    \begin{subfigure}[t]{0.5\textwidth}
        \centering
        \includegraphics[width=0.49\textwidth]{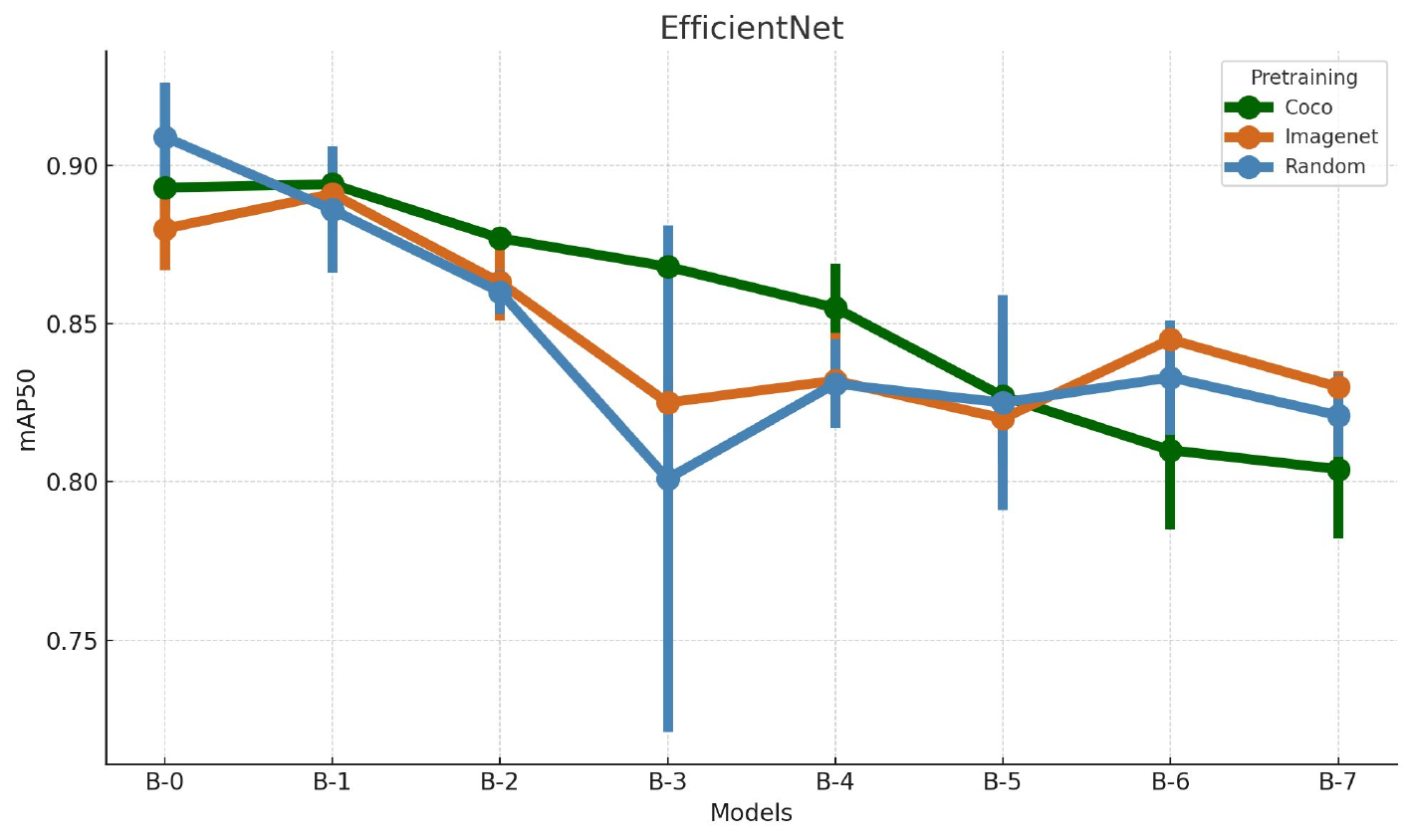}
    \hfill
        \includegraphics[width=0.49\textwidth]{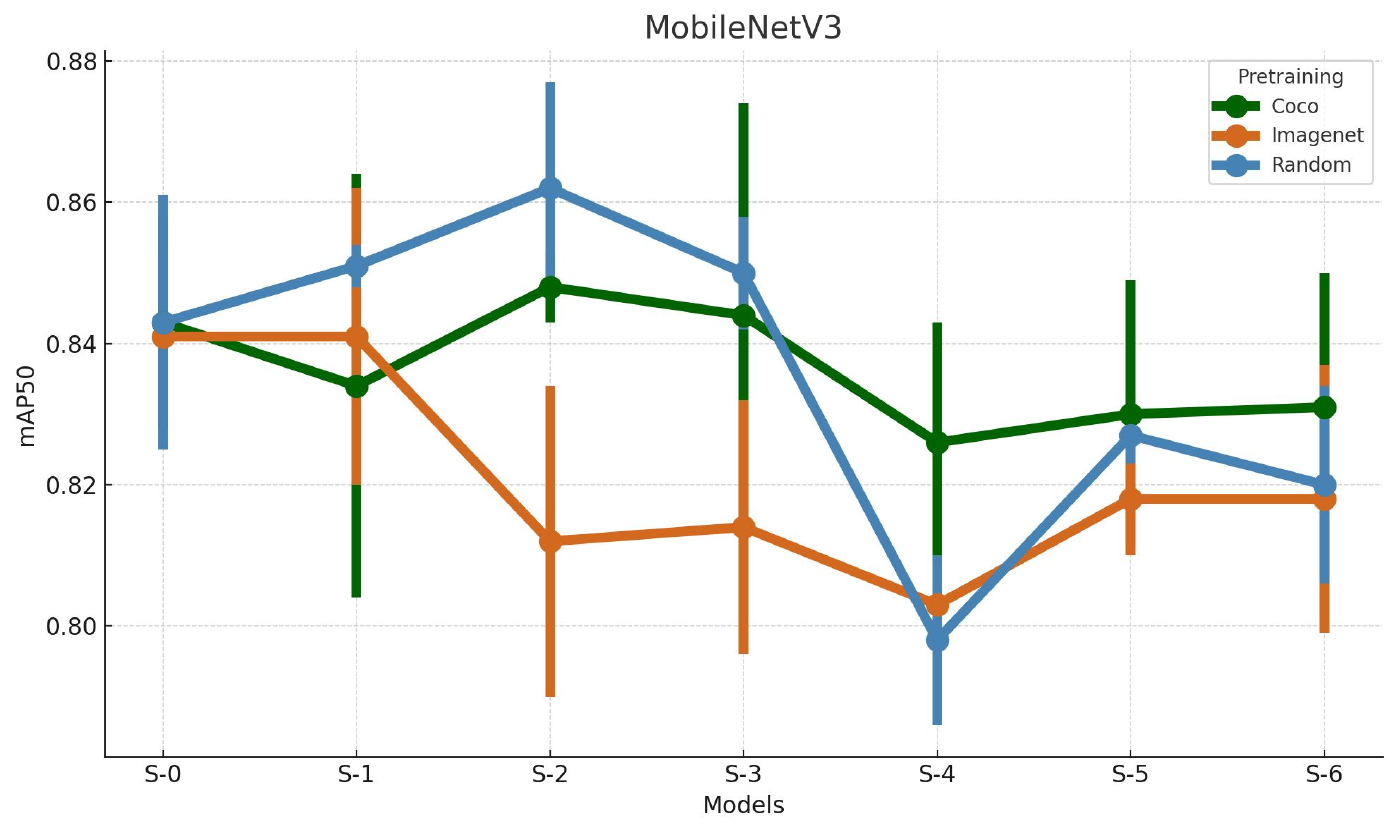}
    \end{subfigure}
    \caption{\textbf{Out-of-Distribution performance on Distech data.} As the domain gap becomes small like in this case, generalization from pretraining is not observed in a consistent way. 
    }
    \label{fig:Distech}
\end{figure}

\paragraph{Viewpoint generalization:} In this experiment we evaluate the generalization performance of our models on Distech data with or without pretraining, when fine-tuning on a set of viewpoints, and testing on other, unseen viewpoints.
As in previous experiments, Fig.~\ref{fig:Distech} reports the generalization performance with our two families of approaches (EfficientNet and MobileNet).
In this case, we cannot see a clear trend in performance when varying the model size. 
We believe this is due to the fact that in this case the domain gap from different point of view is smaller than a change of modality. Thus, when the domain gap is small, the effect of pretraining is not noticeable. A similar trend is also observable on the other datasets for in-domain data, where fine-tuning and evaluation are performed on the same data distribution (see supplementary material). 

\begin{figure}
    \centering
    \begin{subfigure}[t]{0.5\textwidth}
        \centering
        \includegraphics[width=0.49\textwidth]{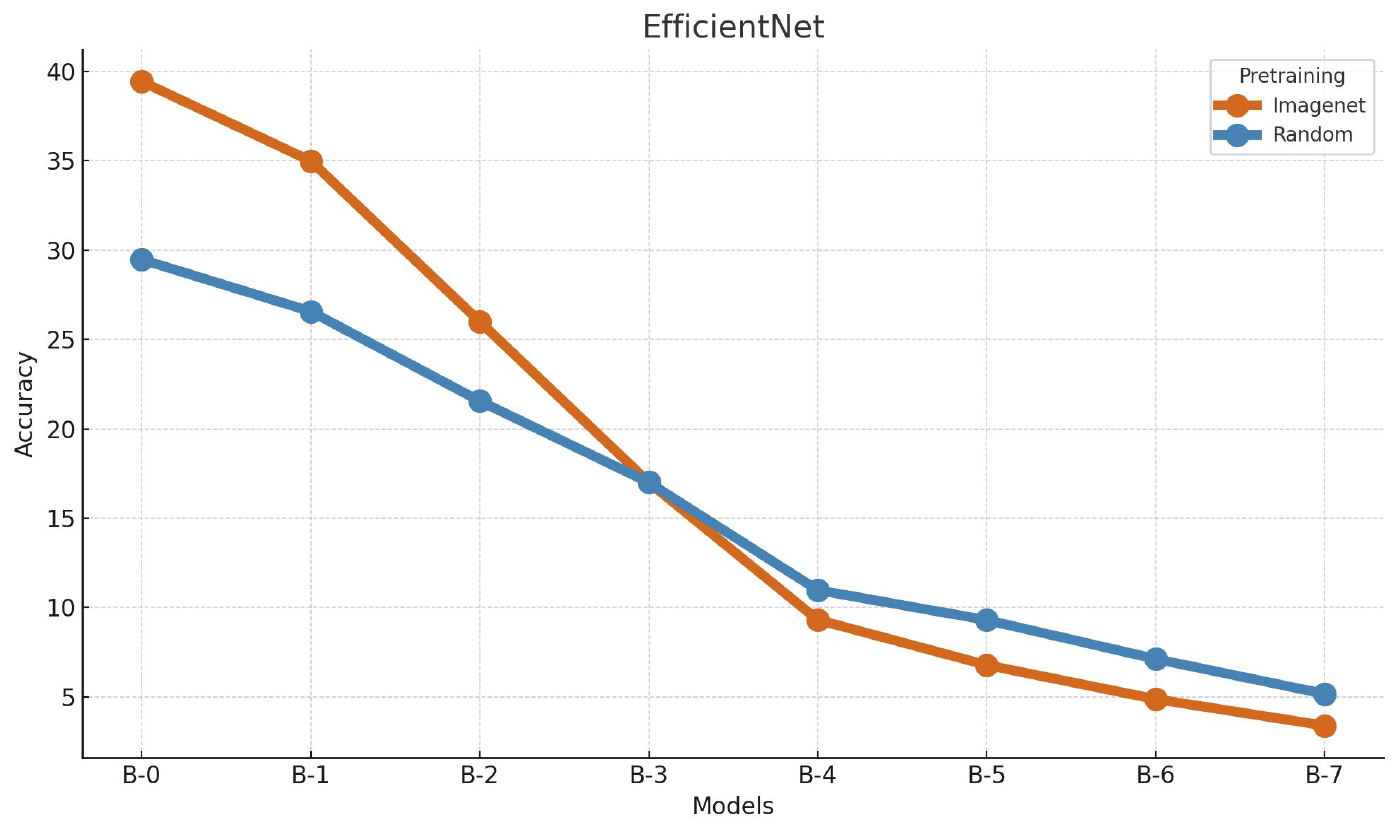}
    \hfill
        \includegraphics[width=0.49\textwidth]{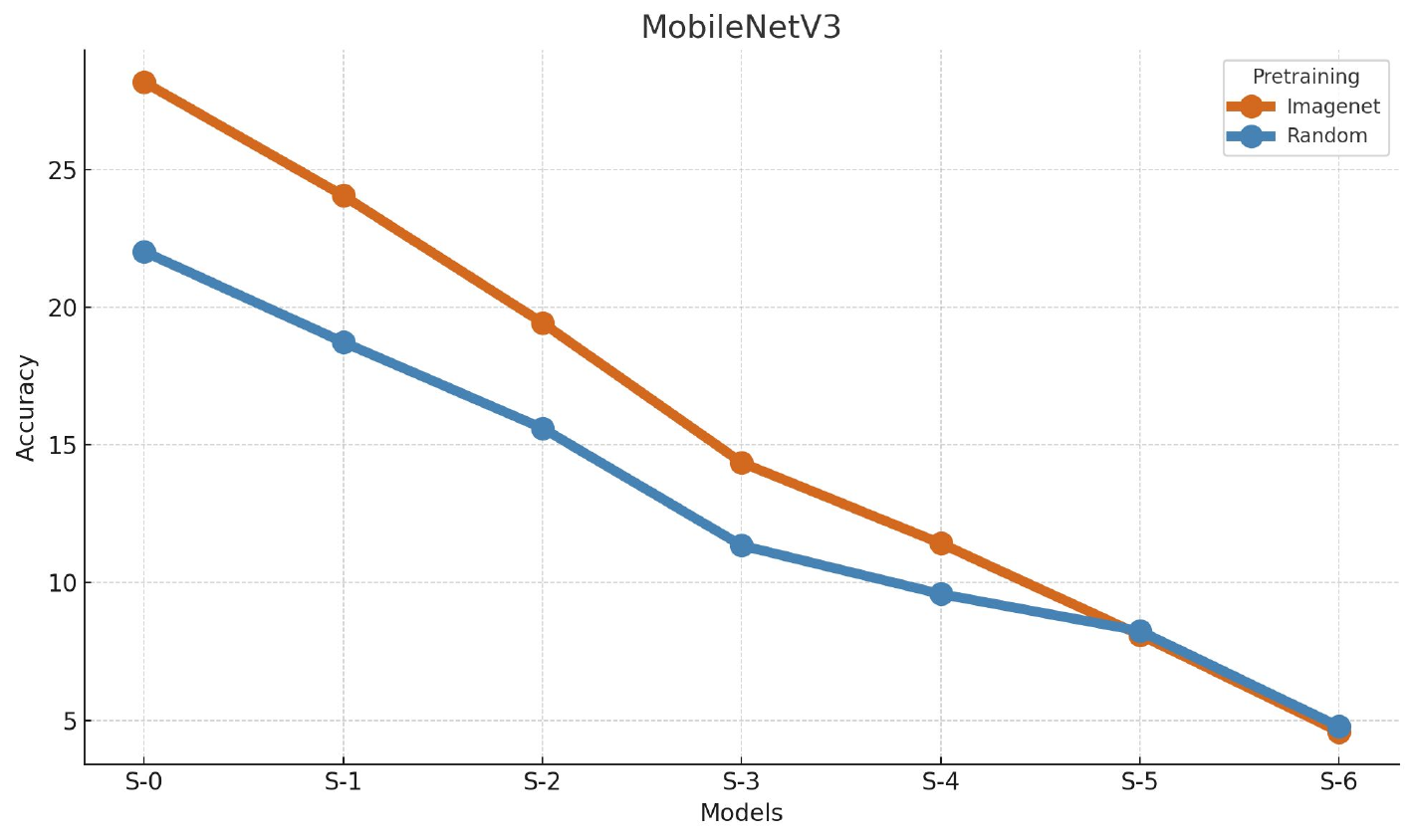}
    \end{subfigure}
    \caption{\textbf{Domain Generalization performance of ultra-small EfficientNet and MobileNet models on DomainNet data averaged across six DomainNet target domains.} Similar to object detection, pretraining helps large as well as certain moderate size backbones.  
    }
    \label{fig:DomainNet}
\end{figure}

\begin{figure}
    \centering
    \hspace{-.5cm}
    \includegraphics[width=0.5\textwidth]{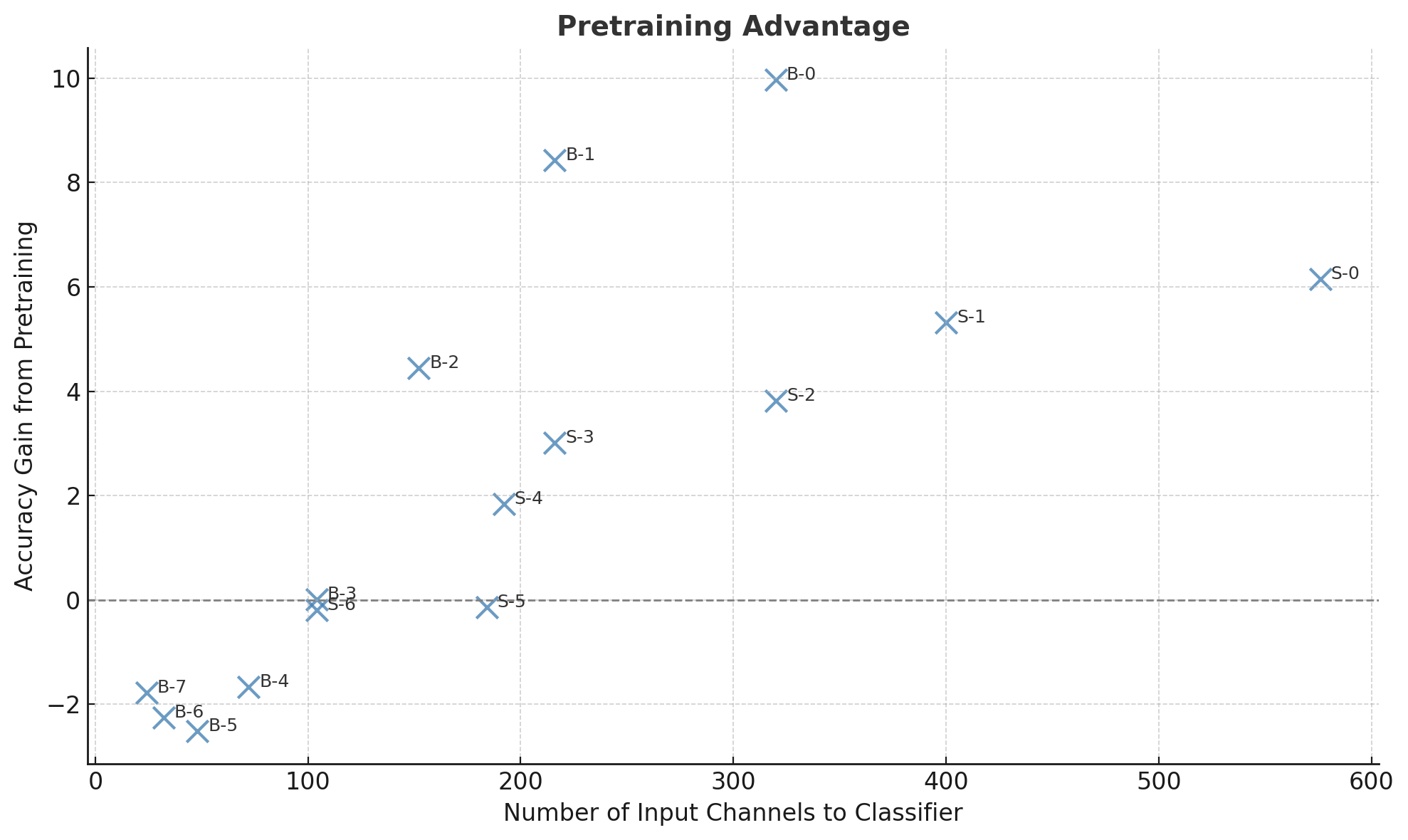}
    \caption{\textbf{Impact of classifier input channel dimension on pretraining gain in domain generalization performance observed on DomainNet benchmark.} The gains obtained from pretraining is strongly correlated with backbone last conv layer channel width which has considerable impact on the capacity of the classification head and in turn the model.
    }
    \label{fig:DomainNet_channel}
\end{figure}

\paragraph{Domain generalization for Image Classification}
Fig.~\ref{fig:DomainNet} shows the results of pretraining and model trained from scratch on the DomainNet benchmark. As shown in the figure, ImageNet pretraining clearly improves accuracy for large as well as certain moderate‑size models in both families (B-0 to B-2 and S-0 to S-4). However, as capacity shrinks, the gap narrows and can invert: the smallest variants (e.g., B‑7, S‑6) show little benefit or a slight edge for scratch. From Fig.~\ref{fig:DomainNet_channel} We also observe impact of classifier channel widths, where pretraining benefit diminishes and even turns negative—once architectures become ultra-small and the collapsed channel widths that is a by-product of this reduction. This suggests that representational bottlenecks, rather than initialization, dominate performance in the smallest backbones. In short, pretraining helps when the model is big enough to leverage it. Random initialization becomes competitive as the model becomes ultra‑small, in line with generalization experiments for object detection discussed in previous sections.


\begin{figure}
    \centering
    \hspace{-.5cm}
    \includegraphics[width=0.5\textwidth]{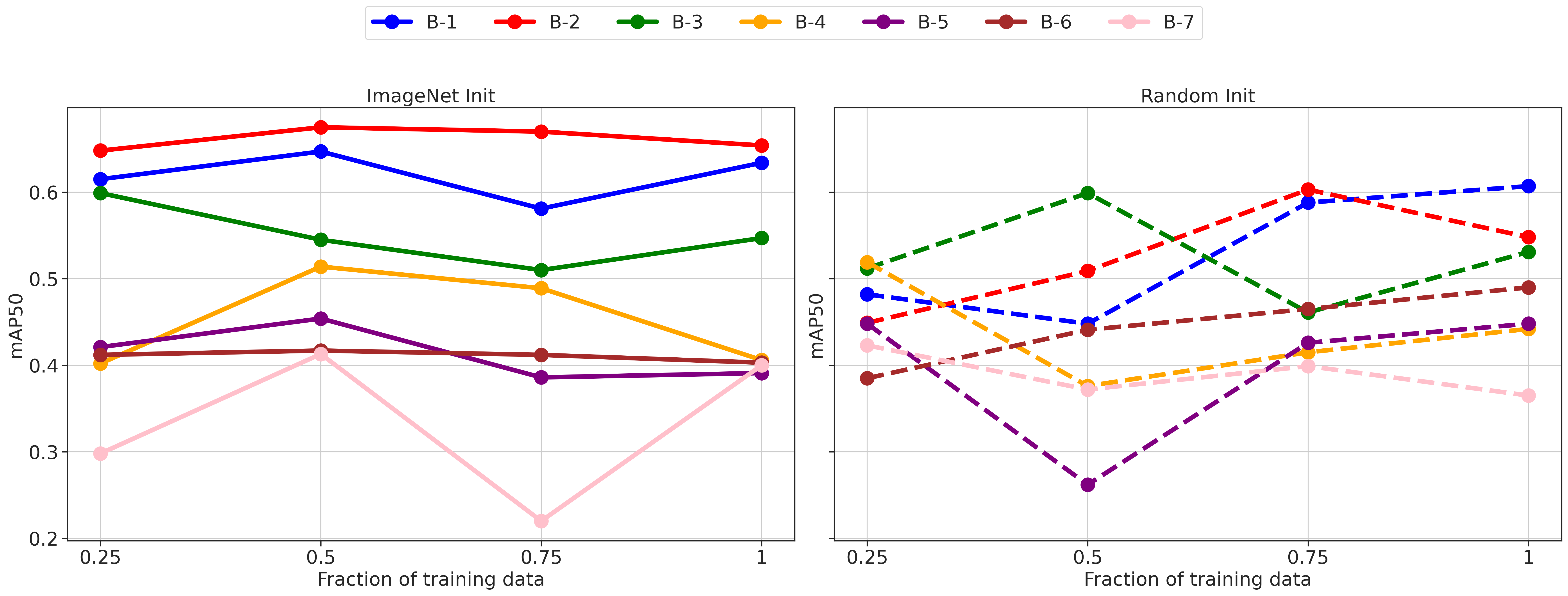}
    \caption{\textbf{Influence of dataset size on out-of-distribution performance.} Out-of-distribution accuracy is shown as a function of the fraction of training data under two initialization strategies: \emph{Imagenet init} (left) and \emph{Random init} (right). Each curve corresponds to a different model variant ($l{=}1$–$7$). Results highlight that increasing the dataset size generally improves performance, though the effect varies depending on initialization and model depth.}

    \label{abl:data}
\end{figure}

\begin{figure*}[h]
\captionsetup[subfigure]{labelformat=empty}
\centering

\textbf{LLVIP - EfficientNet}

\begin{subfigure}[t]{0.50\columnwidth}
    \caption{B-1 - pretrained}
    \makebox[0pt][r]{\makebox[15pt]{\raisebox{50pt}{\rotatebox[origin=c]{90}{ID}}}}%
    \includegraphics[width=\columnwidth]{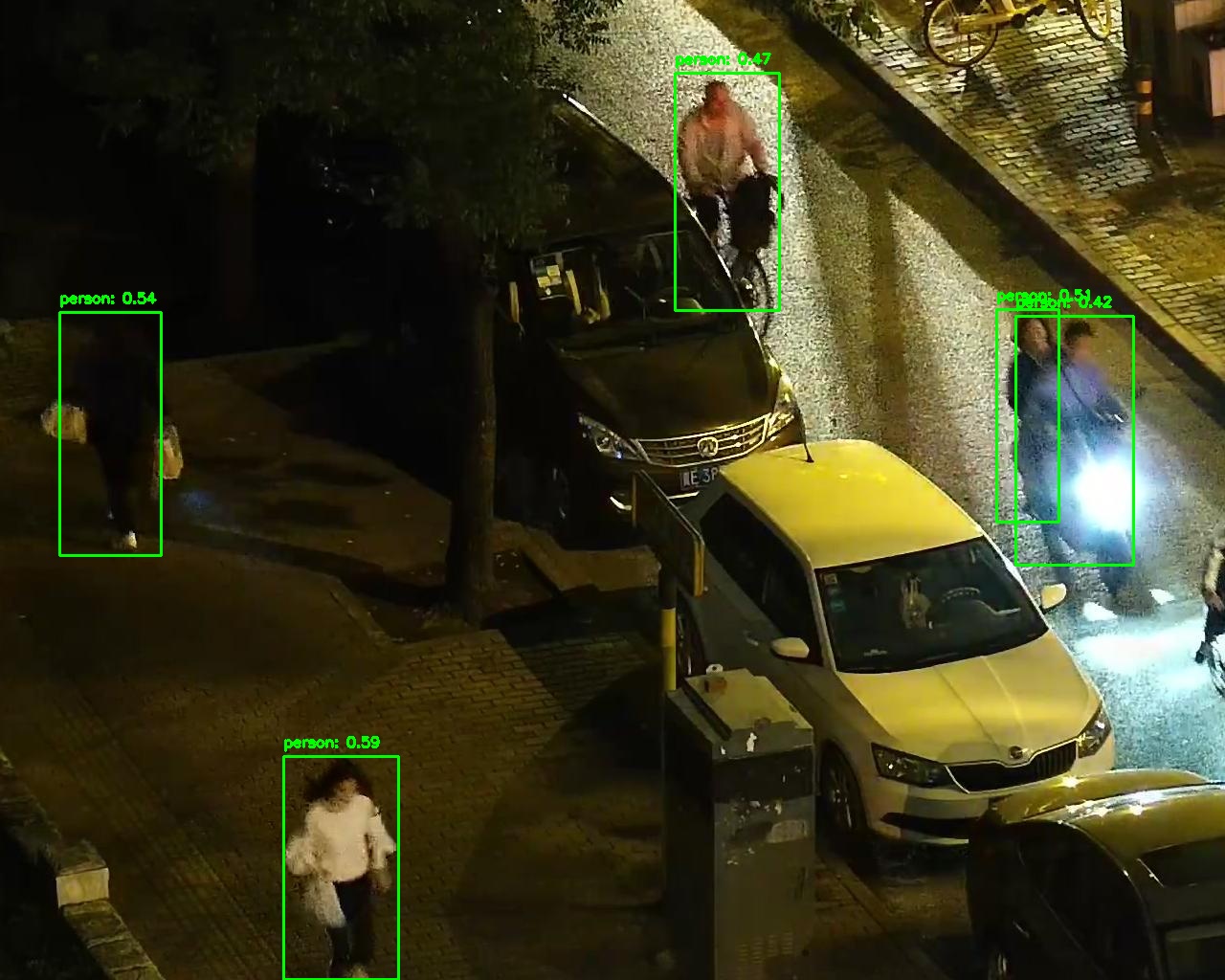}
\end{subfigure}
\begin{subfigure}[t]{0.50\columnwidth}
    \caption{B-3 - pretrained}
    \includegraphics[width=\columnwidth]{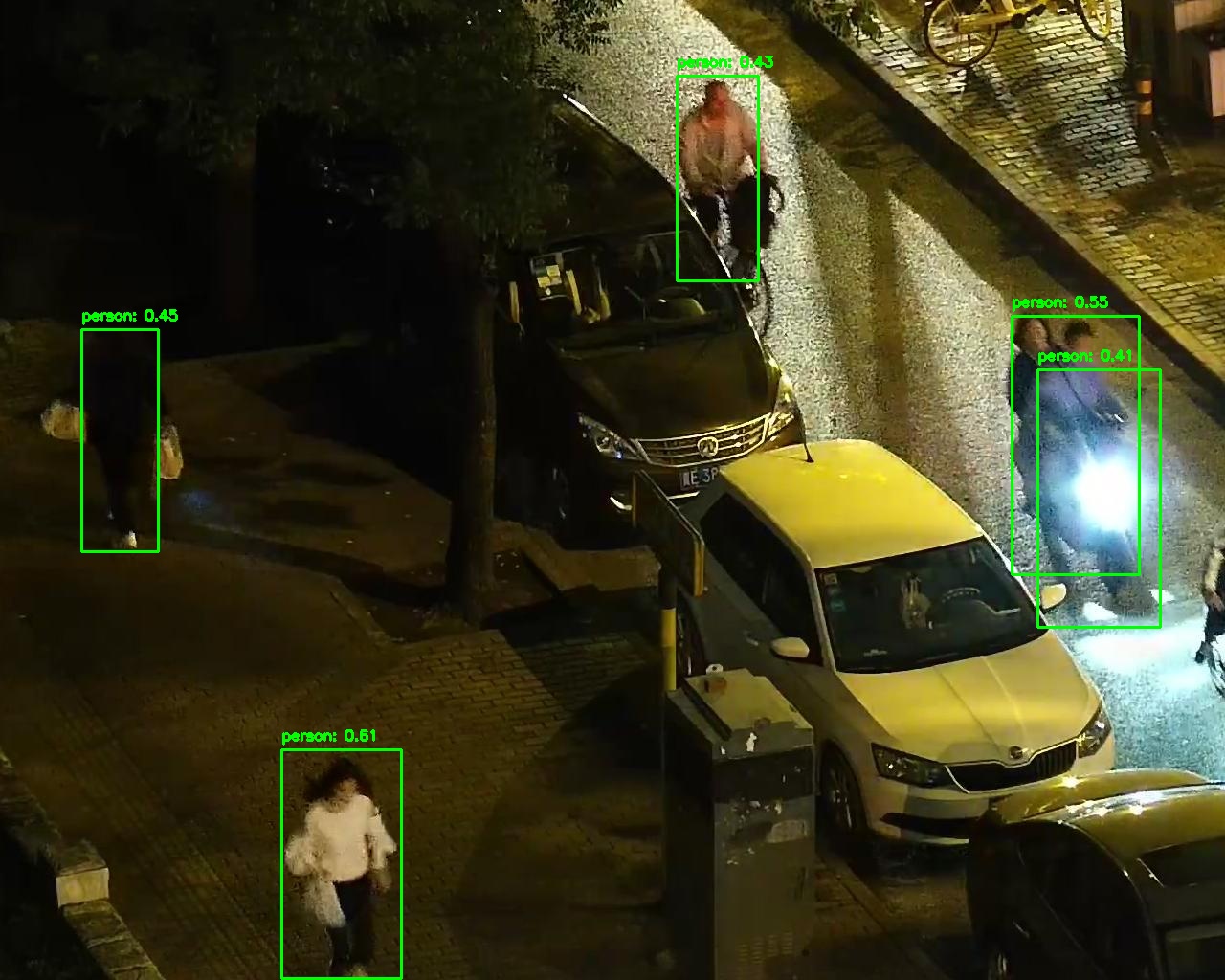}
\end{subfigure}
\begin{subfigure}[t]{0.50\columnwidth}
    \caption{B-1 - Scratch}
    \includegraphics[width=\columnwidth]{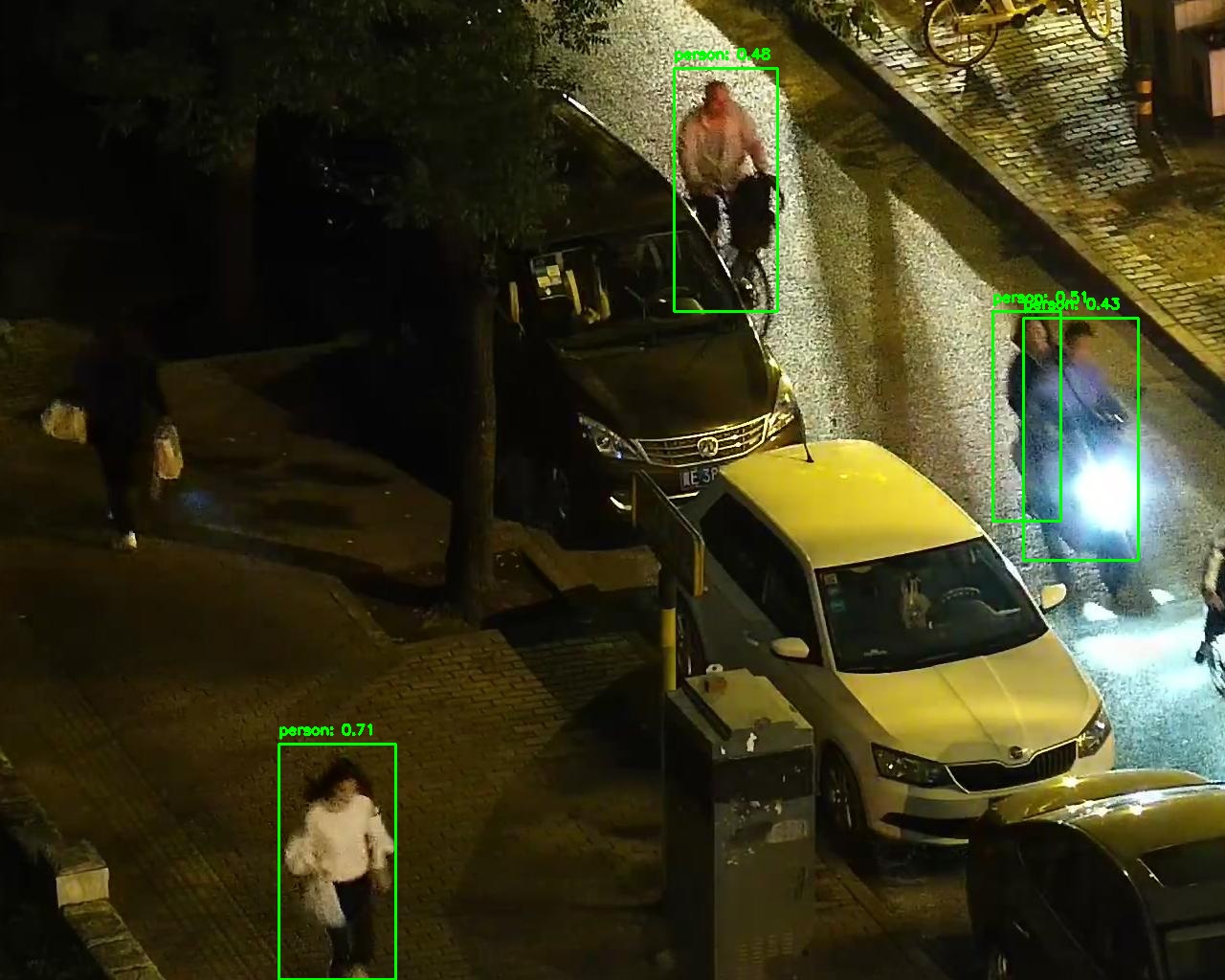}
\end{subfigure}
\begin{subfigure}[t]{0.50\columnwidth}
    \caption{B-3 - Scratch}
    \includegraphics[width=\columnwidth]{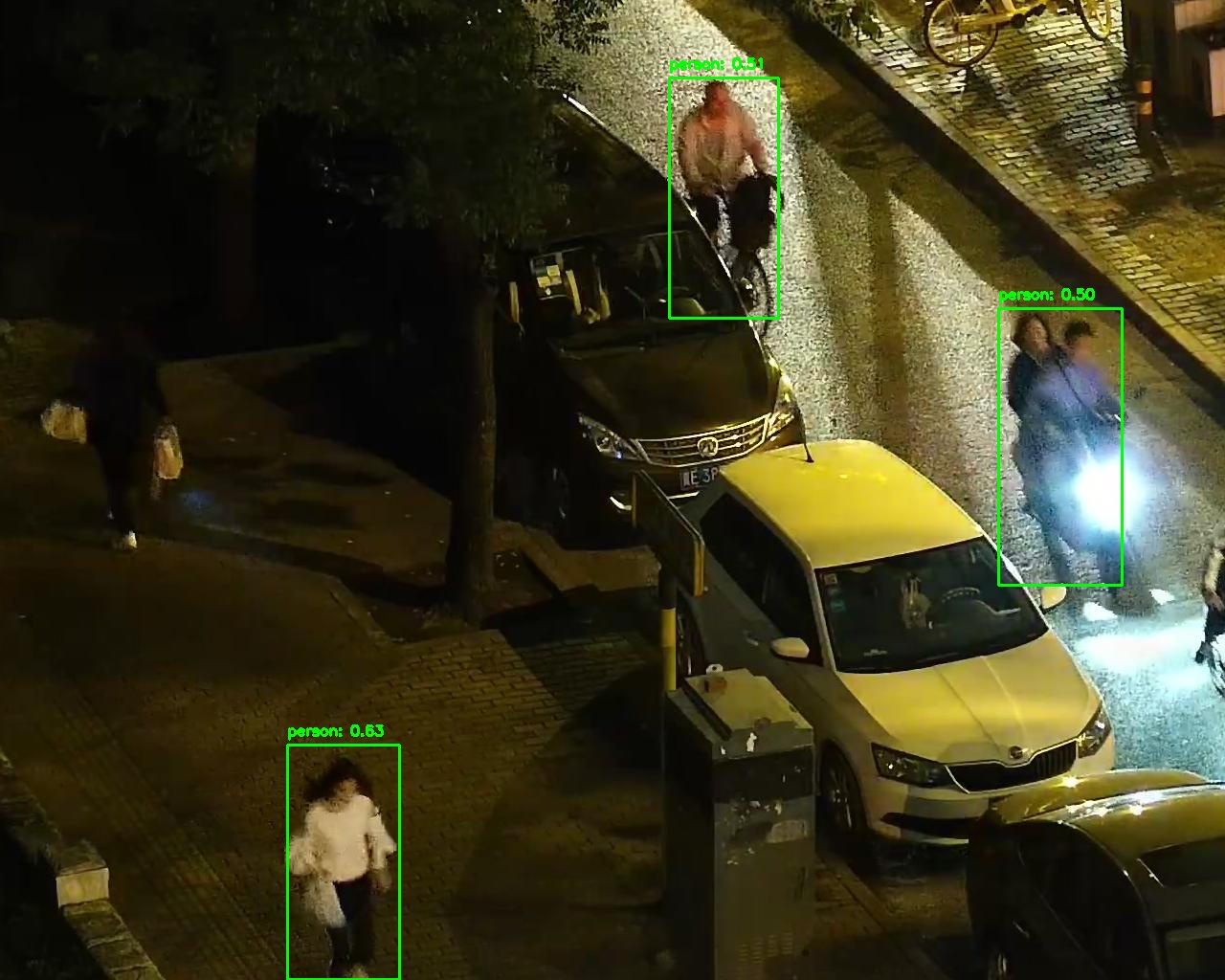}
\end{subfigure}
\begin{subfigure}[t]{0.50\columnwidth}
    \caption{B-1 - pretrained}
    \makebox[0pt][r]{\makebox[15pt]{\raisebox{50pt}{\rotatebox[origin=c]{90}{OOD}}}}%
    \includegraphics[width=\columnwidth]{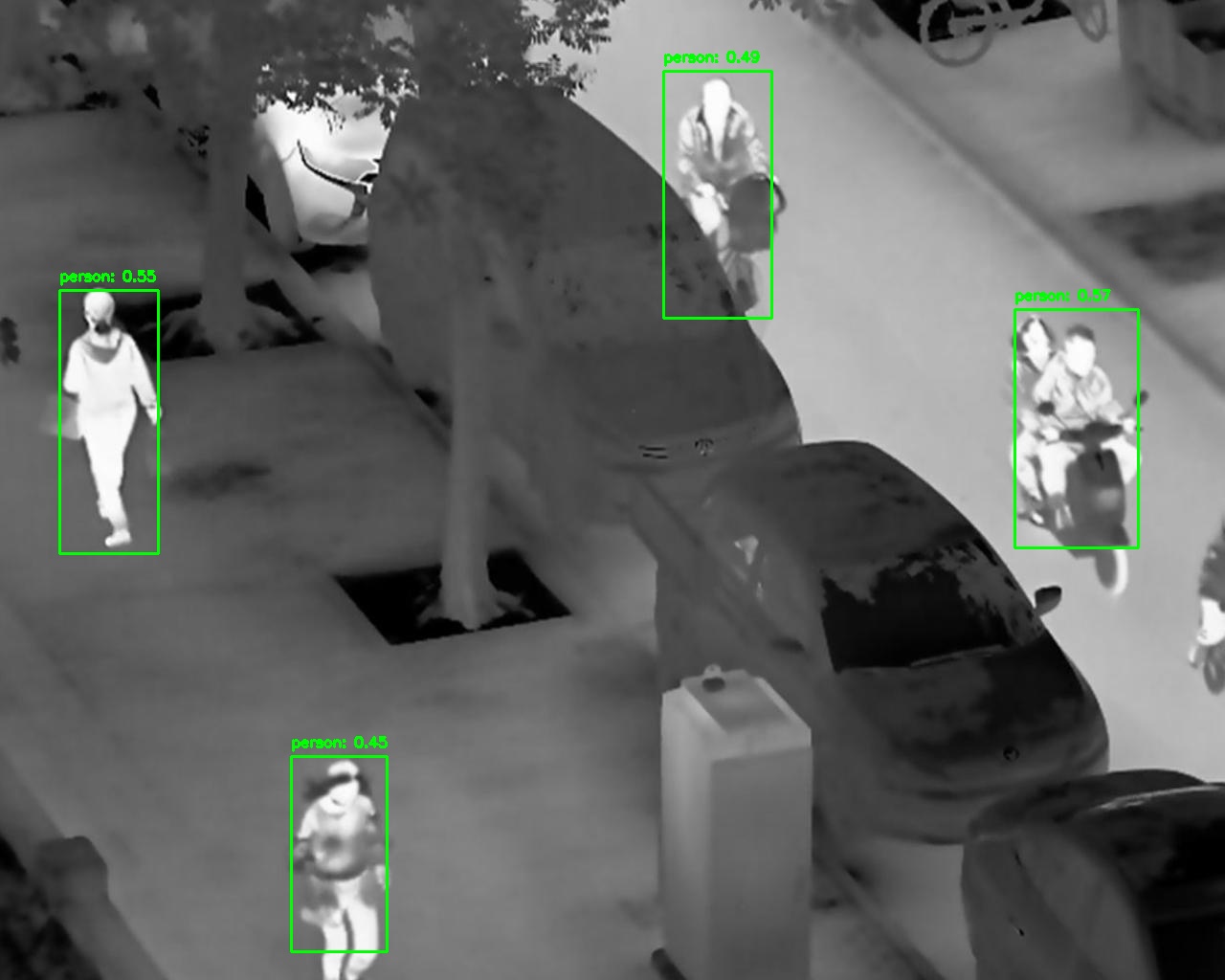}
\end{subfigure}
\begin{subfigure}[t]{0.50\columnwidth}
    \caption{B-3 - pretrained}
    \includegraphics[width=\columnwidth]{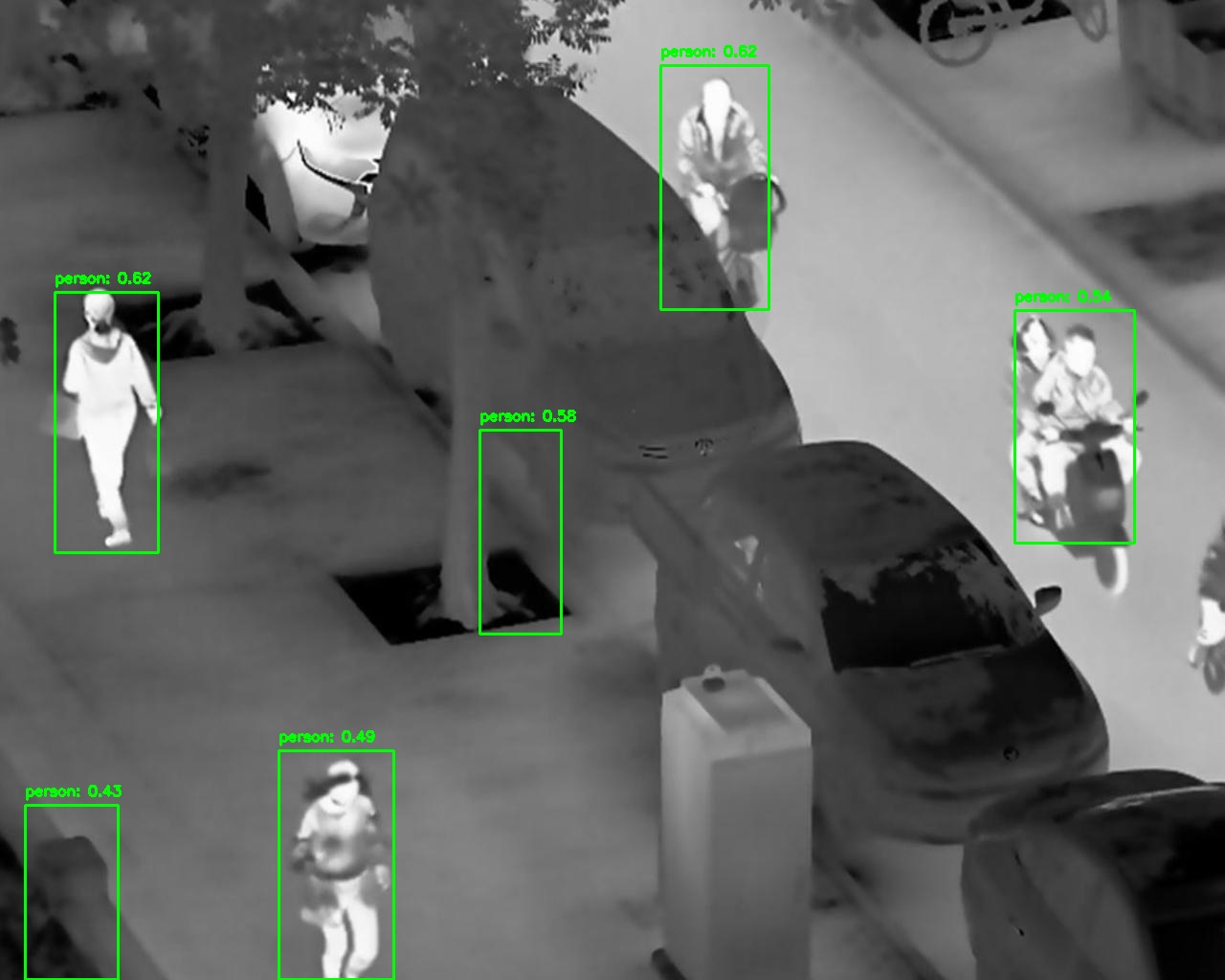}
\end{subfigure}
\begin{subfigure}[t]{0.50\columnwidth}
    \caption{B-1 - Scratch}
    \includegraphics[width=\columnwidth]{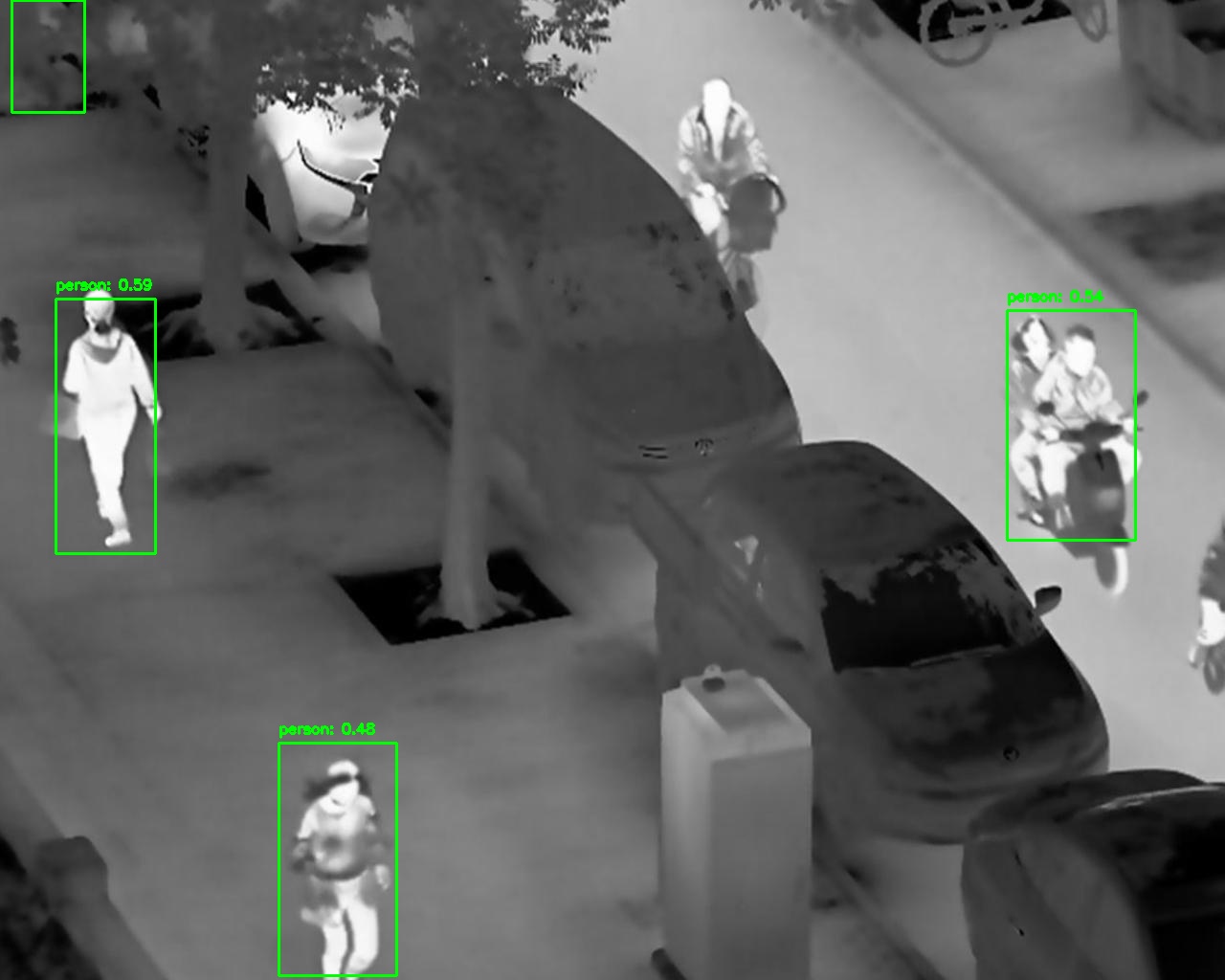}
\end{subfigure}
\begin{subfigure}[t]{0.50\columnwidth}
    \caption{B-3 - Scratch}
    \includegraphics[width=\columnwidth]{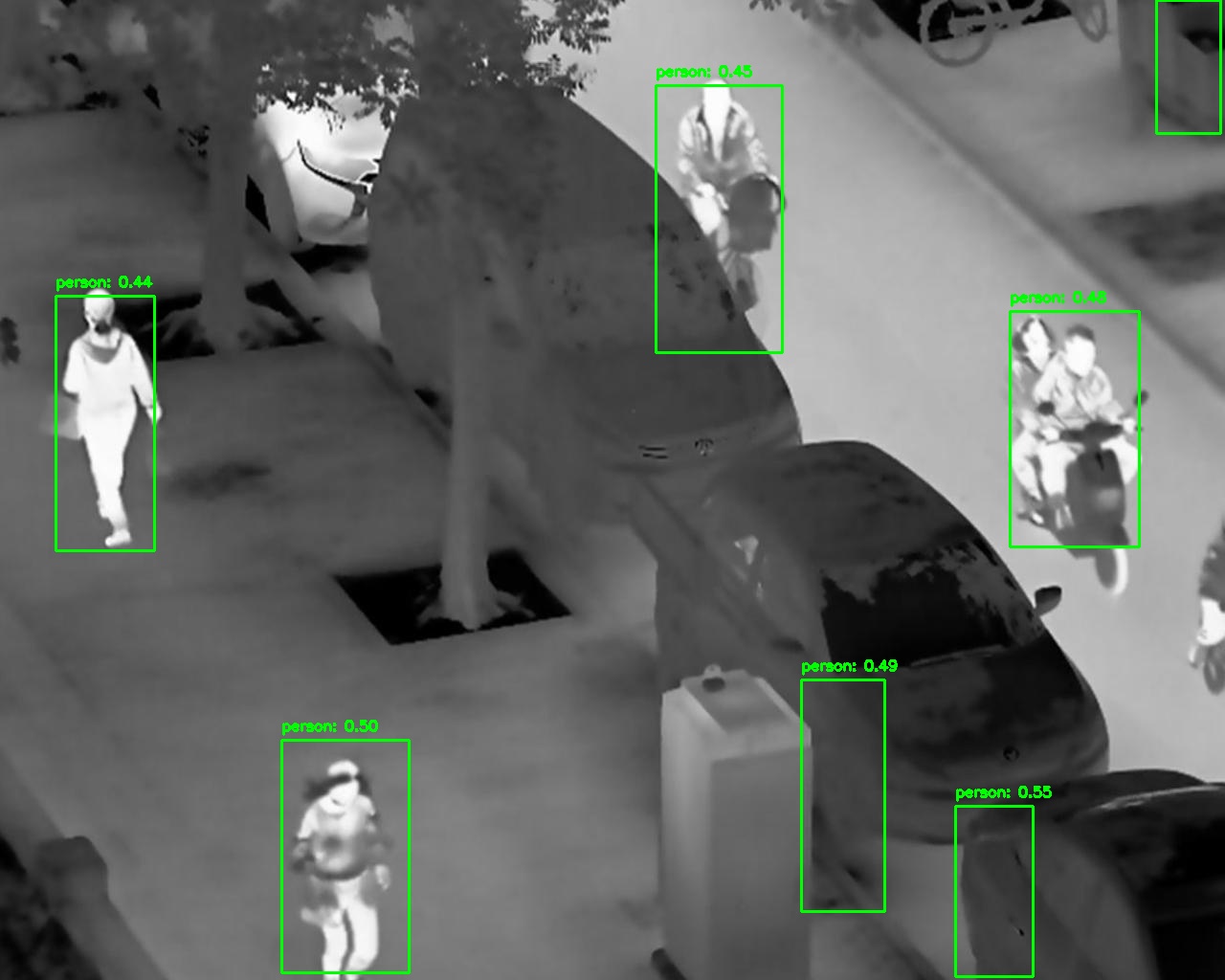}
\end{subfigure}

\caption{\textbf{Qualitative comparison of EfficientNet trained under different settings on the LLVIP dataset.} Here, we present a few interesting qualitative examples for our models. The top row (ID) shows in-distribution RGB samples, while the bottom row (OOD) shows out-of-distribution IR samples. Columns compare pretrained vs. from-scratch models across two different configurations (B-1 and B-3). Green boxes denote detected objects.}
~\label{fig:visualization:LLVIP} .
\end{figure*}

\begin{figure*}[h]
\captionsetup[subfigure]{labelformat=empty}
\centering

\textbf{FLIR - MobileNet}

\begin{subfigure}[t]{0.50\columnwidth}
    \caption{S-1 - pretrained}
    \makebox[0pt][r]{\makebox[15pt]{\raisebox{50pt}{\rotatebox[origin=c]{90}{ID}}}}%
    \includegraphics[width=\columnwidth]{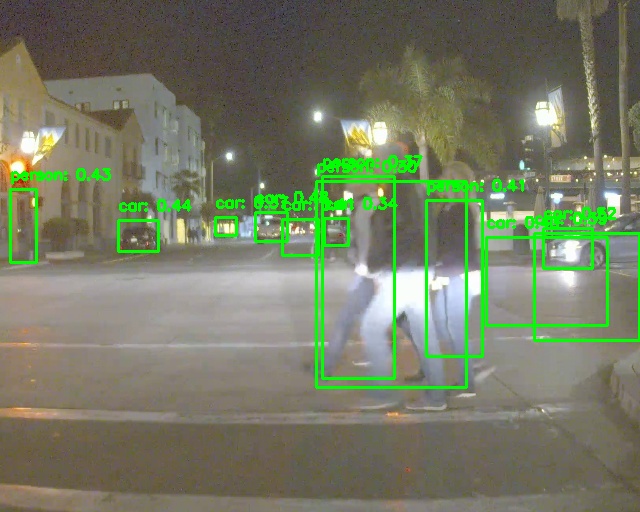}
\end{subfigure}
\begin{subfigure}[t]{0.50\columnwidth}
    \caption{S-3 - pretrained}
    \includegraphics[width=\columnwidth]{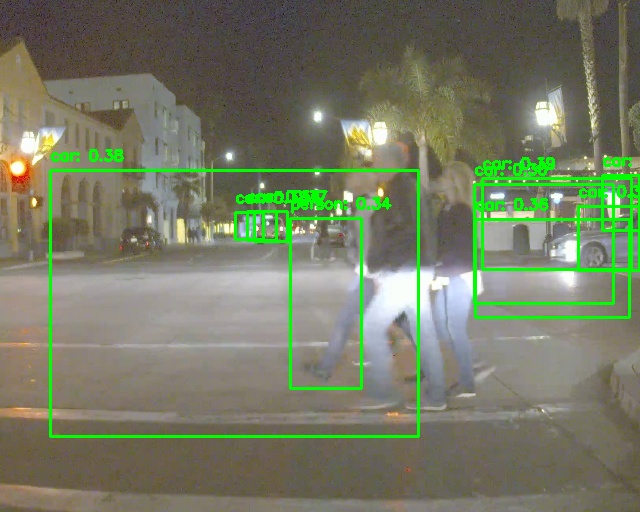}
\end{subfigure}
\begin{subfigure}[t]{0.50\columnwidth}
    \caption{S-1 - Scratch}
    \includegraphics[width=\columnwidth]{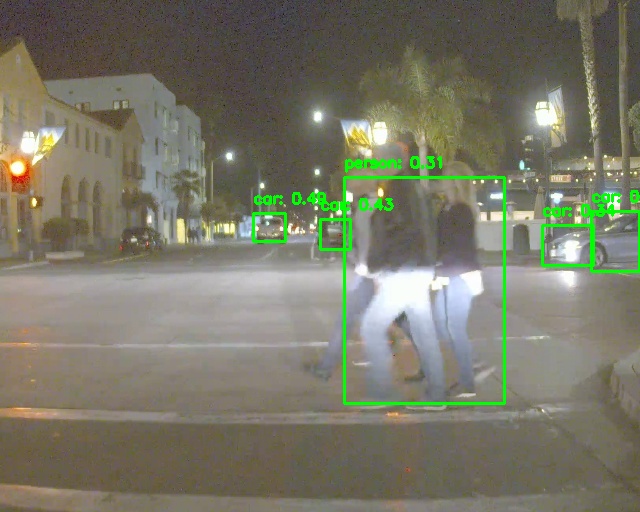}
\end{subfigure}
\begin{subfigure}[t]{0.50\columnwidth}
    \caption{S-3 - Scratch}
    \includegraphics[width=\columnwidth]{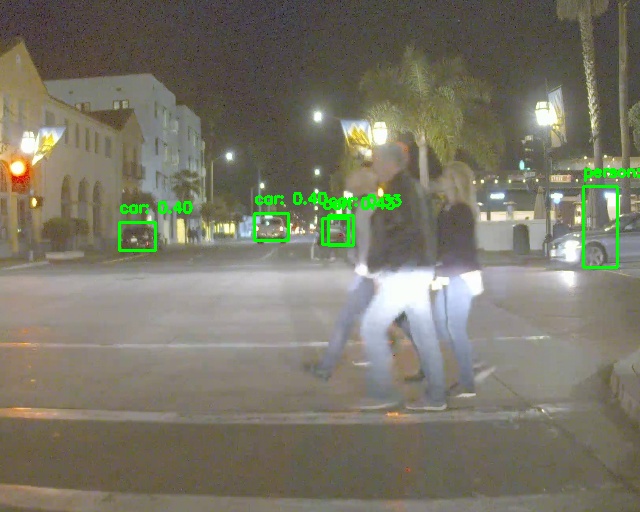}
\end{subfigure}
\begin{subfigure}[t]{0.50\columnwidth}
    \caption{S-1 - pretrained}
    \makebox[0pt][r]{\makebox[15pt]{\raisebox{50pt}{\rotatebox[origin=c]{90}{OOD}}}}%
    \includegraphics[width=\columnwidth]{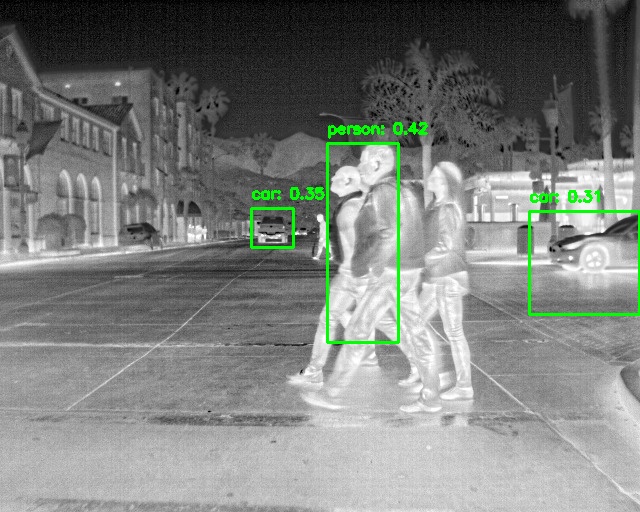}
\end{subfigure}
\begin{subfigure}[t]{0.50\columnwidth}
    \caption{S-3 - pretrained}
    \includegraphics[width=\columnwidth]{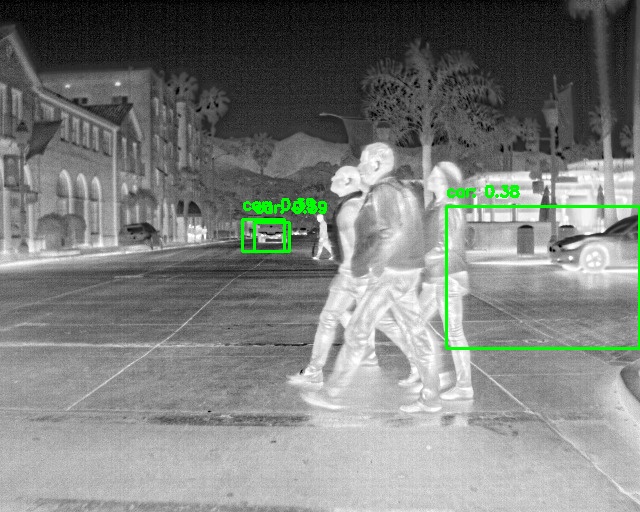}
\end{subfigure}
\begin{subfigure}[t]{0.50\columnwidth}
    \caption{S-1 - Scratch}
    \includegraphics[width=\columnwidth]{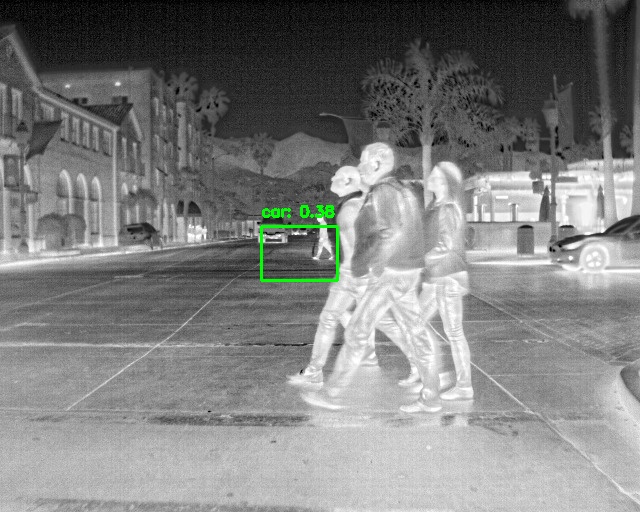}
\end{subfigure}
\begin{subfigure}[t]{0.50\columnwidth}
    \caption{S-3 - Scratch}
    \includegraphics[width=\columnwidth]{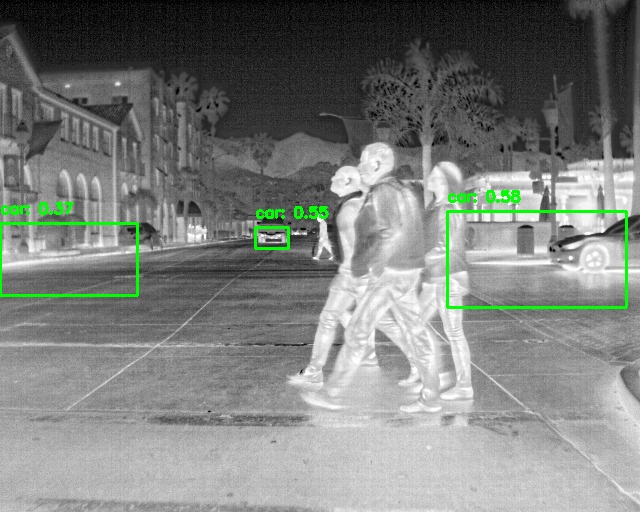}
\end{subfigure}

\caption{\textbf{Qualitative comparison of MobileNet trained under different settings on the FLIR dataset.} The top row (ID) shows in-distribution RGB samples, while the bottom row (OOD) shows out-of-distribution IR samples. Columns compare pretrained vs. from-scratch models across two different configurations (S-1 and S-3). Green boxes denote detected objects. Presence of objects of numerous scale and diversity present challenges in both ID and OOD conditions for both the models.}
\label{fig:visualization:FLIR}
\end{figure*}

\paragraph{Influence of fine-tuning data size} 

Fig.~\ref{abl:data} shows how ImageNet pretraining influences out-of-distribution performance on LLVIP as the x-axis sweeps from 25\% to 100\% of the RGB fine-tuning data, using the EfficientNet backbone family. With ImageNet initialisation (left panel), the two largest variants, B-1 and B-2, outperform their randomly initialised counterparts (right panel) across every data fraction, echoing the pattern seen in Figure 1. The remaining variants paint a subtler picture. pretrained B-3 through B-7 hit their peak mAP50 after seeing roughly half of the RGB data and then level off; the same architectures trained from scratch need the full dataset to approach or occasionally match those peaks.

In conclusion, ImageNet pretraining effectively trades annotation effort for performance: even very small backbones equipped with these weights can rival or surpass larger models trained from scratch. When labels are scarce or deployment budgets are tight, combining a compact backbone with pretraining is the most efficient strategy. Only when plentiful RGB annotations are available does a randomly initialised model begin to close the gap, and even then, it generally falls several points short of its pretrained counterpart.

\paragraph{Detections} Fig.~\ref{fig:visualization:LLVIP} shows qualitative visualizations on LLVIP dataset. We observe EfficientNet model pretrained models show superior detection quality compared to the model trained from scratch. Specifically, We find the pretrained models generally having lower number of false positives compared to the models trained from scratch. Furthermore, larger model B-1, in this example has the most accurate detection with fewest false positives and false negatives compared to smaller pretrained B-3 or other models trained from scratch.  pretrained models generally demonstrate more accurate detections under all conditions. However, both model families face challenges when numerous diverse scaled objects are present as in an example from FLIR dataset in Fig.~\ref{fig:visualization:FLIR}.
\section{Conclusion}

This paper asked a focused question: When does pre‑training help ultra‑small vision models to generalize? Our answer is twofold. First, \textbf{objective alignment matters}: detection‑specific IN$\rightarrow$COCO pretraining paradigm transfers more reliably to cross-domain and cross-modal detection than classification pretraining (ImageNet), particularly for larger backbones and in easier transfer directions. Second, \textbf{capacity gates benefit}: as backbones shrink to the ultra‑small regime, the measurable advantage of pretraining rapidly diminishes and can disappear or reverse for both detection and classification applications. While evaluating SSL was beyond the scope of this work due to the focus on controlled supervised pretraining comparisons, we believe that extending our analysis to SSL-based initializations would be a valuable direction for future work.



\textbf{Acknowledgments}: This work was supported by Distech Controls Inc., the Natural Sciences and Engineering Research Council of Canada, the Digital Research Alliance of Canada, and MITACS.
{
    \small
    \bibliographystyle{ieeenat_fullname}
    \bibliography{main}
}
\clearpage
\twocolumn[{
\begin{center}
\parbox{\textwidth}{\centering
{\Large\bfseries Supplementary Material for\par}
\vspace{0.3em}
{\Large\bfseries Pretraining Helps When Capacity Allows: Evidence from Ultra-Small ConvNets\par}
\vspace{0.8em}
}
\end{center}
}]
\setcounter{section}{0}
\section{Small Model Generation}
\subsection{Smaller variants for EfficientNet}
\label{sec:effnet}

EfficientNet-B0 comprises a stem module, seven subsequent blocks, a convolutional head, and final classification components. In this section, we describe our process of scaling down EfficientNet-B0, which contains millions of parameters, to ultra-small networks with only tens of thousands of parameters.

We adopt two protocols for constructing these compact models. First, we downscale the width and depth of EfficientNet without modifying its block semantics. In the second protocol, we alter the semantics of the EfficientNet blocks by modifying the squeeze-and-excite modules.

\subsubsection{Depth-Width Downscaling of EfficientNet}
\label{effnetdownscale}

We downscale the width and depth of EfficientNet using its original scaling law conventions while maintaining constant input dimensions. Since our end task is object detection, we keep the input resolution unchanged to avoid degrading its performance. Specifically, EfficientNet introduces a compound scaling method that uniformly scales the network depth, width, and resolution using a set of predetermined constants~\cite{tan2019efficientnet}. We modify this compound scaling method to keep the resolution constant, while changing width and depth for our models. Our scaling operation is performed as follows:

\begin{equation}
\begin{aligned}
    \text{Depth}: & \quad d = \alpha^\phi,  \\
    \text{Width}: & \quad w = \beta^\phi, \\
\end{aligned}
\end{equation}

subject to the constraint:

\begin{equation}
\alpha \cdot \beta^2 \approx 2,
\end{equation}

\noindent where $\phi$ is the compound scaling coefficient that controls overall model size, and $\alpha, \beta$ are constants determined via grid search to balance performance and efficiency. It is important to note here that the key distinction between compound scaling used in EfficientNet and ours is the absence of resolution scaling factor $\gamma$. The compound scaling done this way therefore enables us to stay as close to EfficientNet's original compound scaling as possible while retaining focus on our end task which is object detection in non-RGB visual domains. We start from EfficientNet-B0 and scale down until seven levels below it. Therefore the value $\phi$ in our case ranges from $-7$ to $-1$.

\subsection{Smaller variants for MobileNet-V3}
\label{sec:mbnet}
We start from Mobilenet-V3 small~\cite{howard2017mobilenets} and downscale upto six levels after the small model. With each downscaling, for each model, we roughly cut the number of parameters by roughly 50\% compared to the model a level higher to it. We downsize both the width and the depth using the scaling framework from Mobilenet-v3. Further, for each downscaling we use same multiplier parameter for both width and depth multiplier. 
\begin{table*}[ht]
\centering
\begin{tabular}{|l l c c c c|}
\hline
Model & Pretraining & mAP-ID & mAP50-ID & mAP-OOD & mAP50-OOD \\
\hline
B-0 & Coco     & 0.457 (0.004) & 0.854 (0.002) & 0.328 (0.025) & 0.619 (0.049) \\
B-0 & Imagenet & 0.439 (0.000)     & 0.848 (0.000)     & 0.326 (0.000)     & 0.666 (0.000)     \\
B-0 & None     & 0.410 (0.000)     & 0.799 (0.000)     & 0.259 (0.000)     & 0.547 (0.000)     \\
B-1 & Coco     & 0.443 (0.005) & 0.831 (0.005) & 0.330 (0.012) & 0.642 (0.018) \\
B-1 & Imagenet & 0.408 (0.014) & 0.795 (0.024) & 0.304 (0.006) & 0.634 (0.026) \\
B-1 & None     & 0.415 (0.007) & 0.801 (0.011) & 0.300 (0.027) & 0.607 (0.054) \\
B-2 & Coco     & 0.410 (0.006) & 0.792 (0.008) & 0.310 (0.008) & 0.655 (0.014) \\
B-2 & Imagenet & 0.388 (0.010) & 0.777 (0.015) & 0.299 (0.007) & 0.654 (0.023) \\
B-2 & None     & 0.378 (0.007) & 0.752 (0.016) & 0.256 (0.006) & 0.548 (0.019) \\
B-3 & Coco     & 0.367 (0.003) & 0.761 (0.015) & 0.275 (0.002) & 0.617 (0.010) \\
B-3 & Imagenet & 0.360 (0.002) & 0.742 (0.003) & 0.245 (0.020) & 0.547 (0.048) \\
B-3 & None     & 0.358 (0.005) & 0.737 (0.010) & 0.238 (0.006) & 0.531 (0.009) \\
B-4 & Coco     & 0.346 (0.003) & 0.748 (0.009) & 0.189 (0.006) & 0.469 (0.013) \\
B-4 & Imagenet & 0.342 (0.005) & 0.751 (0.012) & 0.146 (0.016) & 0.406 (0.053) \\
B-4 & None     & 0.349 (0.007) & 0.738 (0.012) & 0.187 (0.033) & 0.442 (0.084) \\
B-5 & Coco     & 0.357 (0.003) & 0.760 (0.002) & 0.190 (0.006) & 0.457 (0.011) \\
B-5 & Imagenet & 0.340 (0.003) & 0.750 (0.005) & 0.152 (0.006) & 0.391 (0.012) \\
B-5 & None     & 0.346 (0.018) & 0.730 (0.036) & 0.194 (0.019) & 0.448 (0.045) \\
B-6 & Coco     & 0.328 (0.003) & 0.716 (0.006) & 0.126 (0.013) & 0.322 (0.035) \\
B-6 & Imagenet & 0.326 (0.004) & 0.726 (0.010) & 0.162 (0.020) & 0.403 (0.050) \\
B-6 & None     & 0.339 (0.007) & 0.731 (0.015) & 0.199 (0.020) & 0.490 (0.056) \\
B-7 & Coco     & 0.322 (0.002) & 0.709 (0.009) & 0.204 (0.006) & 0.519 (0.018) \\
B-7 & Imagenet & 0.313 (0.007) & 0.710 (0.012) & 0.152 (0.008) & 0.400 (0.018) \\
B-7 & None     & 0.331 (0.010) & 0.735 (0.009) & 0.151 (0.067) & 0.365 (0.147) \\
\hline
\end{tabular}
\caption{Performance comparison on LLVIP dataset~\cite{jia2021llvip} for EfficientNet family models.}
\label{tab:llvip_effnet}
\end{table*}

\begin{table*}[h]
\centering
\begin{tabular}{|l l c c c c|}
\hline
Model & Pretraining & mAP-ID & mAP50-ID & mAP-OOD & mAP50-OOD \\
\hline
S-0 & Coco     & 0.391 (0.011) & 0.763 (0.019) & 0.263 (0.009) & 0.590 (0.011) \\
S-0 & Imagenet & 0.397 (0.000) & 0.807 (0.000) & 0.272 (0.000) & 0.596 (0.000) \\
S-0 & None     & 0.359 (0.000) & 0.751 (0.000) & 0.237 (0.000) & 0.515 (0.000) \\
S-1 & Coco     & 0.383 (0.001) & 0.763 (0.002) & 0.240 (0.024) & 0.523 (0.054) \\
S-1 & Imagenet & 0.370 (0.005) & 0.757 (0.008) & 0.257 (0.017) & 0.591 (0.019) \\
S-1 & None     & 0.358 (0.011) & 0.733 (0.013) & 0.221 (0.016) & 0.513 (0.027) \\
S-2 & Coco     & 0.377 (0.002) & 0.781 (0.006) & 0.222 (0.026) & 0.521 (0.050) \\
S-2 & Imagenet & 0.364 (0.004) & 0.764 (0.002) & 0.243 (0.005) & 0.559 (0.020) \\
S-2 & None     & 0.351 (0.020) & 0.722 (0.040) & 0.224 (0.031) & 0.521 (0.065) \\
S-3 & Coco     & 0.348 (0.003) & 0.732 (0.013) & 0.207 (0.020) & 0.490 (0.047) \\
S-3 & Imagenet & 0.344 (0.001) & 0.730 (0.004) & 0.189 (0.024) & 0.457 (0.049) \\
S-3 & None     & 0.322 (0.005) & 0.687 (0.011) & 0.222 (0.009) & 0.528 (0.003) \\
S-4 & Coco     & 0.341 (0.003) & 0.731 (0.008) & 0.234 (0.007) & 0.577 (0.009) \\
S-4 & Imagenet & 0.321 (0.005) & 0.709 (0.015) & 0.210 (0.006) & 0.505 (0.011) \\
S-4 & None     & 0.311 (0.009) & 0.671 (0.015) & 0.176 (0.015) & 0.439 (0.026) \\
S-5 & Coco     & 0.343 (0.005) & 0.743 (0.007) & 0.131 (0.008) & 0.320 (0.022) \\
S-5 & Imagenet & 0.317 (0.004) & 0.709 (0.013) & 0.139 (0.027) & 0.340 (0.064) \\
S-5 & None     & 0.329 (0.015) & 0.717 (0.025) & 0.214 (0.006) & 0.519 (0.012) \\
S-6 & Coco     & 0.299 (0.001) & 0.675 (0.005) & 0.214 (0.008) & 0.519 (0.012) \\
S-6 & Imagenet & 0.287 (0.010) & 0.663 (0.026) & 0.211 (0.012) & 0.538 (0.021) \\
S-6 & None     & 0.312 (0.007) & 0.689 (0.014) & 0.180 (0.009) & 0.453 (0.029) \\
\hline
\end{tabular}
\caption{Performance comparison on LLVIP dataset~\cite{jia2021llvip} for MobilenetV3 family models.}
\label{tab:llvip_mbnet}
\end{table*}

\begin{table*}[h]
\centering
\begin{tabular}{|l l c c c c|}
\hline
Model & Pretraining & mAP-ID & mAP50-ID & mAP-OOD & mAP50-OOD \\
\hline
B-0 & Coco     & 0.269 (0.003) & 0.593 (0.005) & 0.147 (0.008) & 0.373 (0.022) \\
B-0 & Imagenet & 0.249 (0.000) & 0.560 (0.000) & 0.122 (0.000) & 0.303 (0.000) \\
B-0 & None     & 0.256 (0.000) & 0.560 (0.000) & 0.113 (0.000) & 0.277 (0.000) \\
B-1 & Coco     & 0.259 (0.002) & 0.568 (0.006) & 0.108 (0.001) & 0.278 (0.004) \\
B-1 & Imagenet & 0.245 (0.001) & 0.545 (0.003) & 0.096 (0.007) & 0.252 (0.023) \\
B-1 & None     & 0.239 (0.000) & 0.521 (0.002) & 0.070 (0.027) & 0.169 (0.072) \\
B-2 & Coco     & 0.226 (0.002) & 0.508 (0.004) & 0.086 (0.004) & 0.223 (0.011) \\
B-2 & Imagenet & 0.214 (0.002) & 0.489 (0.005) & 0.076 (0.007) & 0.190 (0.013) \\
B-2 & None     & 0.203 (0.002) & 0.458 (0.006) & 0.054 (0.027) & 0.137 (0.068) \\
B-3 & Coco     & 0.196 (0.002) & 0.443 (0.003) & 0.052 (0.002) & 0.129 (0.003) \\
B-3 & Imagenet & 0.182 (0.002) & 0.424 (0.006) & 0.044 (0.002) & 0.120 (0.006) \\
B-3 & None     & 0.181 (0.004) & 0.413 (0.012) & 0.043 (0.012) & 0.113 (0.036) \\
B-4 & Coco     & 0.175 (0.001) & 0.402 (0.003) & 0.049 (0.003) & 0.142 (0.006) \\
B-4 & Imagenet & 0.163 (0.000) & 0.381 (0.002) & 0.032 (0.002) & 0.090 (0.004) \\
B-4 & None     & 0.161 (0.003) & 0.371 (0.006) & 0.036 (0.013) & 0.098 (0.034) \\
B-5 & Coco     & 0.184 (0.002) & 0.423 (0.005) & 0.045 (0.002) & 0.117 (0.008) \\
B-5 & Imagenet & 0.168 (0.002) & 0.393 (0.005) & 0.029 (0.005) & 0.075 (0.012) \\
B-5 & None     & 0.162 (0.005) & 0.376 (0.010) & 0.037 (0.010) & 0.103 (0.027) \\
B-6 & Coco     & 0.171 (0.003) & 0.399 (0.006) & 0.033 (0.002) & 0.091 (0.006) \\
B-6 & Imagenet & 0.160 (0.002) & 0.380 (0.006) & 0.030 (0.007) & 0.089 (0.017) \\
B-6 & None     & 0.160 (0.004) & 0.371 (0.006) & 0.028 (0.005) & 0.078 (0.015) \\
B-7 & Coco     & 0.169 (0.001) & 0.394 (0.004) & 0.036 (0.003) & 0.104 (0.010) \\
B-7 & Imagenet & 0.157 (0.001) & 0.371 (0.005) & 0.026 (0.003) & 0.069 (0.007) \\
B-7 & None     & 0.159 (0.005) & 0.369 (0.010) & 0.036 (0.010) & 0.099 (0.027) \\
\hline
\end{tabular}
\caption{Performance comparison on FLIR dataset\cite{zhang2020multispectral} for EfficientNet family models.}
\label{tab:flir:EfficientNet}
\end{table*}

\begin{table*}[h]
\centering
\begin{tabular}{|l l c c c c|}
\hline
Model & Pretraining & mAP-ID & mAP50-ID & mAP-OOD & mAP50-OOD \\
\hline
S-0 & Coco     & 0.216 (0.000) & 0.485 (0.000) & 0.088 (0.000) & 0.225 (0.000) \\
S-0 & Imagenet & 0.202 (0.001) & 0.463 (0.002) & 0.059 (0.003) & 0.157 (0.006) \\
S-0 & None     & 0.184 (0.003) & 0.416 (0.002) & 0.043 (0.006) & 0.118 (0.014) \\
S-1 & Coco     & 0.195 (0.001) & 0.447 (0.004) & 0.071 (0.006) & 0.186 (0.018) \\
S-1 & Imagenet & 0.181 (0.003) & 0.419 (0.004) & 0.055 (0.003) & 0.155 (0.014) \\
S-1 & None     & 0.177 (0.002) & 0.404 (0.007) & 0.039 (0.005) & 0.107 (0.015) \\
S-2 & Coco     & 0.182 (0.002) & 0.416 (0.000) & 0.052 (0.002) & 0.143 (0.003) \\
S-2 & Imagenet & 0.172 (0.003) & 0.395 (0.005) & 0.047 (0.001) & 0.127 (0.002) \\
S-2 & None     & 0.174 (0.003) & 0.393 (0.006) & 0.036 (0.007) & 0.101 (0.024) \\
S-3 & Coco     & 0.164 (0.001) & 0.378 (0.003) & 0.050 (0.003) & 0.141 (0.010) \\
S-3 & Imagenet & 0.156 (0.003) & 0.358 (0.004) & 0.023 (0.004) & 0.067 (0.009) \\
S-3 & None     & 0.163 (0.003) & 0.373 (0.008) & 0.036 (0.005) & 0.102 (0.013) \\
S-4 & Coco     & 0.164 (0.001) & 0.381 (0.003) & 0.037 (0.001) & 0.121 (0.003) \\
S-4 & Imagenet & 0.150 (0.001) & 0.354 (0.004) & 0.031 (0.004) & 0.090 (0.012) \\
S-4 & None     & 0.150 (0.007) & 0.348 (0.014) & 0.040 (0.003) & 0.115 (0.007) \\
S-5 & Coco     & 0.159 (0.001) & 0.373 (0.002) & 0.030 (0.001) & 0.085 (0.004) \\
S-5 & Imagenet & 0.141 (0.001) & 0.338 (0.004) & 0.022 (0.002) & 0.067 (0.007) \\
S-5 & None     & 0.141 (0.001) & 0.322 (0.004) & 0.035 (0.005) & 0.101 (0.014) \\
S-6 & Coco     & 0.143 (0.001) & 0.339 (0.004) & 0.040 (0.002) & 0.114 (0.004) \\
S-6 & Imagenet & 0.136 (0.002) & 0.323 (0.004) & 0.023 (0.003) & 0.069 (0.005) \\
S-6 & None     & 0.140 (0.002) & 0.330 (0.005) & 0.023 (0.003) & 0.066 (0.007) \\
\hline
\end{tabular}
\caption{Performance comparison on FLIR dataset\cite{zhang2020multispectral} for MobileNetV3 family models.}
\label{tab:flir:Mobilenet}
\end{table*}

\begin{table*}[h]
\centering
\begin{tabular}{|l l c c c c|}
\hline
Model & Pretraining & mAP-ID & mAP50-ID & mAP-OOD & mAP50-OOD \\
\hline
B-0 & Coco     & 0.705 (0.002) & 0.948 (0.001) & 0.663 (0.004) & 0.893 (0.007) \\
B-0 & Imagenet & 0.709 (0.002) & 0.950 (0.001) & 0.655 (0.011) & 0.880 (0.013) \\
B-0 & None     & 0.698 (0.003) & 0.943 (0.003) & 0.665 (0.010) & 0.909 (0.017) \\
B-1 & Coco     & 0.711 (0.003) & 0.947 (0.001) & 0.662 (0.004) & 0.894 (0.007) \\
B-1 & Imagenet & 0.711 (0.001) & 0.947 (0.000) & 0.657 (0.008) & 0.891 (0.010) \\
B-1 & None     & 0.691 (0.003) & 0.941 (0.001) & 0.645 (0.015) & 0.886 (0.020) \\
B-2 & Coco     & 0.629 (0.003) & 0.923 (0.000) & 0.607 (0.002) & 0.877 (0.002) \\
B-2 & Imagenet & 0.625 (0.001) & 0.922 (0.002) & 0.600 (0.010) & 0.863 (0.012) \\
B-2 & None     & 0.610 (0.001) & 0.906 (0.005) & 0.596 (0.006) & 0.860 (0.007) \\
B-3 & Coco     & 0.584 (0.011) & 0.891 (0.008) & 0.581 (0.004) & 0.868 (0.003) \\
B-3 & Imagenet & 0.582 (0.003) & 0.894 (0.004) & 0.555 (0.008) & 0.825 (0.008) \\
B-3 & None     & 0.524 (0.089) & 0.839 (0.072) & 0.525 (0.077) & 0.801 (0.080) \\
B-4 & Coco     & 0.554 (0.003) & 0.873 (0.003) & 0.563 (0.013) & 0.855 (0.014) \\
B-4 & Imagenet & 0.546 (0.003) & 0.871 (0.000) & 0.545 (0.006) & 0.832 (0.015) \\
B-4 & None     & 0.558 (0.002) & 0.874 (0.000) & 0.556 (0.008) & 0.831 (0.014) \\
B-5 & Coco     & 0.557 (0.002) & 0.879 (0.000) & 0.544 (0.012) & 0.827 (0.021) \\
B-5 & Imagenet & 0.550 (0.006) & 0.869 (0.007) & 0.541 (0.002) & 0.820 (0.006) \\
B-5 & None     & 0.563 (0.006) & 0.868 (0.005) & 0.554 (0.025) & 0.825 (0.034) \\
B-6 & Coco     & 0.559 (0.004) & 0.872 (0.004) & 0.540 (0.015) & 0.810 (0.025) \\
B-6 & Imagenet & 0.549 (0.001) & 0.865 (0.003) & 0.560 (0.003) & 0.845 (0.002) \\
B-6 & None     & 0.562 (0.002) & 0.876 (0.000) & 0.553 (0.011) & 0.833 (0.018) \\
B-7 & Coco     & 0.561 (0.002) & 0.870 (0.001) & 0.539 (0.016) & 0.804 (0.022) \\
B-7 & Imagenet & 0.562 (0.000) & 0.875 (0.003) & 0.548 (0.004) & 0.830 (0.005) \\
B-7 & None     & 0.536 (0.039) & 0.856 (0.029) & 0.537 (0.022) & 0.821 (0.013) \\
\hline
\end{tabular}
\caption{Performance comparison on LLVIP dataset~\cite{jia2021llvip} for EfficientNet family models.}
\label{tab:Distech:EfficientNet}
\end{table*}

\begin{table*}[h]
\centering
\begin{tabular}{|l l c c c c|}
\hline
Model & Pretraining & mAP-ID & mAP50-ID & mAP-OOD & mAP50-OOD \\
\hline
S-0 & Coco     & 0.620 (0.007) & 0.911 (0.002) & 0.584 (0.009) & 0.843 (0.015) \\
S-0 & Imagenet & 0.631 (0.004) & 0.923 (0.004) & 0.580 (0.007) & 0.841 (0.012) \\
S-0 & None     & 0.615 (0.002) & 0.910 (0.005) & 0.575 (0.012) & 0.843 (0.018) \\
S-1 & Coco     & 0.611 (0.003) & 0.910 (0.002) & 0.571 (0.024) & 0.834 (0.030) \\
S-1 & Imagenet & 0.609 (0.003) & 0.907 (0.004) & 0.578 (0.016) & 0.841 (0.021) \\
S-1 & None     & 0.594 (0.001) & 0.894 (0.003) & 0.579 (0.002) & 0.851 (0.003) \\
S-2 & Coco     & 0.607 (0.002) & 0.905 (0.001) & 0.578 (0.004) & 0.848 (0.005) \\
S-2 & Imagenet & 0.607 (0.003) & 0.906 (0.002) & 0.556 (0.017) & 0.812 (0.022) \\
S-2 & None     & 0.603 (0.008) & 0.900 (0.002) & 0.589 (0.009) & 0.862 (0.015) \\
S-3 & Coco     & 0.587 (0.001) & 0.896 (0.002) & 0.566 (0.018) & 0.844 (0.030) \\
S-3 & Imagenet & 0.586 (0.003) & 0.896 (0.001) & 0.550 (0.011) & 0.814 (0.018) \\
S-3 & None     & 0.582 (0.003) & 0.893 (0.001) & 0.575 (0.002) & 0.850 (0.008) \\
S-4 & Coco     & 0.539 (0.000) & 0.863 (0.006) & 0.540 (0.013) & 0.826 (0.017) \\
S-4 & Imagenet & 0.546 (0.004) & 0.865 (0.005) & 0.522 (0.004) & 0.803 (0.001) \\
S-4 & None     & 0.540 (0.006) & 0.856 (0.005) & 0.519 (0.010) & 0.798 (0.012) \\
S-5 & Coco     & 0.535 (0.002) & 0.861 (0.004) & 0.540 (0.015) & 0.830 (0.019) \\
S-5 & Imagenet & 0.543 (0.008) & 0.867 (0.007) & 0.537 (0.004) & 0.818 (0.008) \\
S-5 & None     & 0.542 (0.004) & 0.858 (0.002) & 0.545 (0.003) & 0.827 (0.004) \\
S-6 & Coco     & 0.541 (0.005) & 0.865 (0.008) & 0.542 (0.012) & 0.831 (0.019) \\
S-6 & Imagenet & 0.542 (0.002) & 0.860 (0.001) & 0.535 (0.007) & 0.818 (0.019) \\
S-6 & None     & 0.545 (0.008) & 0.860 (0.006) & 0.544 (0.011) & 0.820 (0.014) \\
\hline
\end{tabular}
\caption{Performance comparison on Distech dataset for MobileNetV3 family models.}
\label{tab:Distech:Mobilenet}
\end{table*}

\begin{table}[h]
\centering
\begin{tabular}{|l l c c|}
\hline
Model & Pretraining & mAP-OOD & mAP50-OOD \\
\hline
B-0 & Coco     & 32.930 (4.330) & 12.230 (1.580) \\
B-0 & Imagenet & 21.050 (0.550) &  7.800 (0.100) \\
B-0 & Coco     & 32.370 (5.340) & 12.600 (1.850) \\
B-1 & Imagenet & 23.870 (3.790) &  8.230 (1.430) \\
B-1 & None     &  4.900 (2.660) &  1.870 (1.020) \\
B-1 & Coco     & 21.630 (3.270) &  8.370 (0.980) \\
B-2 & Imagenet & 19.330 (2.160) &  6.800 (0.940) \\
B-2 & None     &  2.130 (2.050) &  0.900 (0.700) \\
B-2 & Coco     & 26.930 (1.730) &  9.530 (0.590) \\
B-3 & Imagenet & 10.170 (1.320) &  3.130 (0.210) \\
B-3 & None     &  1.500 (1.770) &  0.530 (0.610) \\
B-3 & Coco     &  8.400 (6.520) &  2.830 (2.110) \\
B-4 & Imagenet &  3.630 (1.440) &  1.030 (0.450) \\
B-4 & None     &  0.600 (0.780) &  0.170 (0.240) \\
B-4 & Coco     & 18.600 (2.870) &  6.200 (1.350) \\
B-5 & Imagenet &  9.000 (0.830) &  2.900 (0.220) \\
B-5 & None     &  2.100 (1.000) &  0.830 (0.120) \\
B-5 & Coco     &  7.630 (1.130) &  2.500 (0.360) \\
B-6 & Imagenet &  7.270 (1.960) &  2.230 (0.760) \\
B-6 & None     &  0.470 (0.330) &  0.130 (0.120) \\
B-6 & Coco     & 13.430 (2.490) &  4.770 (1.010) \\
B-7 & Imagenet &  7.130 (2.380) &  2.400 (0.820) \\
B-7 & None     &  0.330 (0.340) &  0.070 (0.090) \\
\hline
\end{tabular}
\caption{Performance comparison on cross-dataset FLIR$\rightarrow$LLVIP RGB on person class for EfficientNet family models.}
\label{tab:flir_llvip:EfficientNet}
\end{table}

\begin{table*}[h]
\centering
\begin{tabular}{|l l c c|}
\hline
Model & Pretraining & mAP-OOD & mAP50-OOD \\
\hline
S-0 & Imagenet & 12.300 (4.580) & 4.250 (1.710) \\
S-0 & None     &  3.500 (3.350) & 1.170 (1.020) \\
S-1 & Coco     & 24.930 (3.970) & 9.500 (1.680) \\
S-1 & Imagenet & 17.970 (3.450) & 6.170 (1.390) \\
S-1 & None     &  0.470 (0.330) & 0.100 (0.080) \\
S-2 & Coco     & 19.170 (1.350) & 6.970 (0.410) \\
S-2 & Imagenet & 13.700 (2.550) & 4.670 (0.970) \\
S-2 & None     &  0.600 (0.450) & 0.270 (0.190) \\
S-3 & Coco     & 13.500 (2.890) & 4.670 (1.110) \\
S-3 & Imagenet &  8.470 (3.920) & 2.700 (1.180) \\
S-3 & None     &  0.900 (0.370) & 0.370 (0.190) \\
S-4 & Coco     & 14.470 (2.400) & 5.170 (0.760) \\
S-4 & Imagenet & 12.800 (3.680) & 4.000 (0.860) \\
S-4 & None     &  1.070 (0.450) & 0.300 (0.080) \\
S-5 & Coco     & 11.300 (0.880) & 3.830 (0.310) \\
S-5 & Imagenet &  2.830 (3.020) & 0.930 (1.040) \\
S-5 & None     &  1.030 (0.540) & 0.300 (0.280) \\
S-6 & Coco     & 13.270 (2.520) & 4.530 (0.920) \\
S-6 & Imagenet &  2.230 (0.900) & 0.670 (0.260) \\
S-6 & None     &  1.400 (0.290) & 0.370 (0.170) \\
\hline
\end{tabular}
\caption{Cross-domain performance for LLVIP$\rightarrow$FLIR RGB on person class for MobileNetV3 family models.}
\label{tab:flir_llvip:MobileNet}
\end{table*}

\begin{table*}[h]
\centering
\begin{tabular}{|l l c c|}
\hline
Model & Pretraining & mAP-OOD & mAP50-OOD \\
\hline
B-0 & Coco     & 9.730 (1.170) & 3.430 (0.380) \\
B-0 & Imagenet & 7.130 (0.710) & 2.500 (0.280) \\
B-0 & None     & 4.800 (1.420) & 1.830 (0.390) \\
B-1 & Coco     & 8.030 (0.050) & 2.770 (0.050) \\
B-1 & Imagenet & 6.830 (1.470) & 2.600 (0.510) \\
B-1 & None     & 5.170 (0.260) & 1.970 (0.120) \\
B-2 & Coco     & 6.870 (1.170) & 2.570 (0.450) \\
B-2 & Imagenet & 4.530 (0.450) & 1.670 (0.210) \\
B-2 & None     & 4.130 (0.840) & 1.530 (0.450) \\
B-3 & Coco     & 3.700 (0.330) & 1.400 (0.220) \\
B-3 & Imagenet & 3.600 (0.850) & 1.270 (0.250) \\
B-3 & None     & 2.870 (0.340) & 1.100 (0.140) \\
B-4 & Coco     & 3.700 (0.490) & 1.400 (0.080) \\
B-4 & Imagenet & 3.330 (0.630) & 1.200 (0.290) \\
B-4 & None     & 2.170 (0.340) & 0.970 (0.190) \\
B-5 & Coco     & 2.630 (0.120) & 1.070 (0.050) \\
B-5 & Imagenet & 3.130 (0.330) & 1.170 (0.050) \\
B-5 & None     & 1.830 (1.300) & 0.700 (0.510) \\
B-6 & Coco     & 2.770 (0.120) & 1.000 (0.000) \\
B-6 & Imagenet & 2.300 (0.370) & 0.930 (0.050) \\
B-6 & None     & 2.900 (0.410) & 1.170 (0.120) \\
B-7 & Coco     & 2.130 (0.520) & 0.730 (0.250) \\
B-7 & Imagenet & 2.230 (0.340) & 1.030 (0.050) \\
B-7 & None     & 1.970 (0.540) & 0.870 (0.260) \\
\hline
\end{tabular}
\caption{Performance comparison on cross-dataset LLVIP$\rightarrow$FLIR RGB on person class for EfficientNet family models.}
\label{tab:llvip_flir:EfficientNet}
\end{table*}

\begin{table*}[h]
\centering
\begin{tabular}{|l l c c|}
\hline
Model & Pretraining & mAP-OOD & mAP50-OOD \\
\hline
S-0 & Coco     & 4.530 (0.250) & 2.000 (0.140) \\
S-0 & Imagenet & 3.950 (0.650) & 1.450 (0.250) \\
S-0 & None     & 2.970 (1.190) & 1.030 (0.460) \\
S-1 & Coco     & 4.230 (0.540) & 1.770 (0.190) \\
S-1 & Imagenet & 3.200 (0.400) & 1.400 (0.200) \\
S-1 & None     & 3.270 (0.560) & 1.200 (0.140) \\
S-2 & Coco     & 4.300 (0.080) & 1.570 (0.050) \\
S-2 & Imagenet & 3.770 (0.480) & 1.470 (0.260) \\
S-2 & None     & 3.030 (0.120) & 1.130 (0.050) \\
S-3 & Coco     & 2.700 (0.160) & 1.130 (0.050) \\
S-3 & Imagenet & 2.630 (0.170) & 1.070 (0.050) \\
S-3 & None     & 2.430 (0.400) & 1.000 (0.160) \\
S-4 & Coco     & 3.500 (0.570) & 1.300 (0.360) \\
S-4 & Imagenet & 3.170 (0.210) & 1.230 (0.120) \\
S-4 & None     & 1.770 (0.980) & 0.670 (0.400) \\
S-5 & Coco     & 3.070 (0.390) & 1.230 (0.090) \\
S-5 & Imagenet & 2.750 (0.150) & 1.100 (0.100) \\
S-5 & None     & 2.150 (0.050) & 0.800 (0.100) \\
S-6 & Coco     & 2.070 (0.450) & 0.800 (0.140) \\
S-6 & Imagenet & 2.700 (0.240) & 1.100 (0.080) \\
S-6 & None     & 2.170 (0.170) & 0.830 (0.210) \\
\hline
\end{tabular}
\caption{Performance comparison on cross-dataset LLVIP$\rightarrow$FLIR RGB on person class for MobileNetV3 family models.}
\label{tab:llvip_flir:MobileNet}
\end{table*}

\begin{table*}[h]
\centering
\begin{tabular}{|l l c|}
\hline
Model & Pretraining & Accuracy \\
\hline
B-0 & Imagenet & 39.439 \\
B-1 & Imagenet & 34.993 \\
B-2 & Imagenet & 25.999 \\
B-3 & Imagenet & 17.002 \\
B-4 & Imagenet &  9.303 \\
B-5 & Imagenet &  6.772 \\
B-6 & Imagenet &  4.864 \\
B-7 & Imagenet &  3.381 \\
B-0 & None     & 29.474 \\
B-1 & None     & 26.568 \\
B-2 & None     & 21.553 \\
B-3 & None     & 16.995 \\
B-4 & None     & 10.968 \\
B-5 & None     &  9.289 \\
B-6 & None     &  7.114 \\
B-7 & None     &  5.157 \\
\hline
\end{tabular}
\caption{Performance comparison on DomainNet~\cite{peng2019moment} benchmark for EfficientNet family models.}
\label{tab:domainnet:effnet}
\end{table*}

\begin{table}[h]
\centering
\begin{tabular}{|l l c|}
\hline
Model & Pretraining & Accuracy \\
\hline
S-0 & Imagenet & 28.181 \\
S-1 & Imagenet & 24.059 \\
S-2 & Imagenet & 19.437 \\
S-3 & Imagenet & 14.363 \\
S-4 & Imagenet & 11.427 \\
S-5 & Imagenet &  8.104 \\
S-6 & Imagenet &  4.586 \\
S-0 & None     & 22.025 \\
S-1 & None     & 18.735 \\
S-2 & None     & 15.613 \\
S-3 & None     & 11.349 \\
S-4 & None     &  9.588 \\
S-5 & None     &  8.247 \\
S-6 & None     &  4.773 \\
\hline
\end{tabular}
\caption{Performance comparison on DomainNet~\cite{peng2019moment} benchmark for EfficientNet family models.}
\label{tab:domainnet:mbnet}
\end{table}

\section{Pre-training details}

\noindent \textbf{(b) ImageNet Pre-training Details:} For the ImageNet pretraining, we pretrain our EfficientNet model families as well as MobileNetV3 model families using FFCV framework~\cite{leclerc2022ffcv}. We train for $32$ epochs with base learning rate of $0.05$ and use input resolution of $384\times384$  and a total batch size of $392$. We use SGD optimizer, with the weight decay of $0.0001$ and momentum of $0.9$. We use cyclic scheduling and keep resolution fixed throughout the training. We train our pre-training models at fixed resolutions throught training. The augmentation pipeline comprises a random resized crop, followed by a random horizontal flip with probability of $0.5$ for data augmentation.\\

\noindent \textbf{(b) COCO Pre-training Details:} For COCO detection pretraining, we adopt the single-stage FCOS style object detector from Nanodet~\cite{lyu2021nanodet}. We train for $32$ epochs with base learning rate of $0.005$ and use input resolution of $480\times384$  and a total batch size of $150$. We adopt a PANet~\cite{liu2018path} style feature pyramid network (FPN) as the neck with an output channel dimension of 96 operating on three output scales. For the detection head, we use NanoDetHead~\cite{lyu2021nanodet}, consisting of two stacked convolutional layers with ReLU6 activation and batch normalization. The head uses an input and feature channel dimension of 96, operates with strides (s) of s=[8, 16, 32], and shares classification and regression features. It predicts bounding boxes with a maximum regression bin index (reg\_max) of $7$, using an octave base scale of $5$ with $1$ scale per octave. The loss function is composed of a Quality Focal Loss~\cite{li2020generalized} for classification ($\beta=2.0$, weight=1.0), a Distribution Focal Loss~\cite{li2020generalized} for bounding box distribution regression (weight=0.25), and a GIoU Loss~\cite{rezatofighi2019generalized} for bounding box localization (weight=2.0). We use SGD optimizer, with the weight decay of $0.0001$ and momentum of $0.9$. We use cyclic scheduling and keep resolution fixed throughout the training. We train our pre-training models at fixed resolutions throught training. The augmentation pipeline comprises a random resized crop, followed by a random horizontal flip with probability of $0.5$ for data augmentation. 
\noindent For data augmentation, we apply horizontal flip with a probability of $0.5$, translation with a probability of $0.2$, random scaling between $0.5$ and $1.5$, apply color augmentations. We train our models with batch size of $150$,  use AdamW~\cite{loshchilov2017decoupled} optimizer with linear warmup for 500 steps and CosineAnnealing~\cite{loshchilov2016sgdr} learning rate scheduler. We train Object detector in LLVIP for $32$ epochs and in FLIR for $80$ epochs. We use base learning rate of 0.0005. \\

\section{Pre-training performance of Models}
\subsection{Imagenet classification}
The top-1 and top-5 accuracy of all our models in our two model families at the end of this training is shown in Tab.~\ref{tab:imagenet}.

\begin{table}[ht]
\centering
\label{tab:imagenet_accuracy}
\begin{tabular}{|l c c|}
\hline
\textbf{Model} & \textbf{Top-1 Acc.} & \textbf{Top-5 Acc.} \\
\hline
\multicolumn{3}{c}{\textbf{EfficientNet}} \\
\hline
B-0 & 0.673 & 0.882 \\
B-1  & 0.635 & 0.857 \\
B-2  & 0.554 & 0.798 \\
B-3  & 0.461 & 0.720 \\
B-4  & 0.354 & 0.605 \\
B-5  & 0.316 & 0.563 \\
B-6  & 0.259 & 0.497 \\
B-7  & 0.210 & 0.426 \\
\hline
\multicolumn{3}{c}{\textbf{MobileNetV3}} \\
\hline
S  & 0.55
& 0.795	
\\
S-1  & 0.507  & 0.757 \\
S-2  & 0.450  & 0.711 \\
S-3  & 0.391  & 0.651 \\
S-4  & 0.360  & 0.613 \\
S-5  & 0.323  & 0.572 \\
S-6  & 0.249  & 0.476 \\
\hline
\end{tabular}
\caption{\textbf{ImageNet top-1 and top-5 accuracy for EfficientNet and MobileNetV3 model families at input resolution 384$\times$384 except for EfficientNet-B0 and MobileNet-V3 S}.}
\label{tab:imagenet}
\end{table}

\subsection{COCO Object Detection}
The mAP and mAP50 of all our models in our two model families at the end of COCO detection pre-training is shown in Tab.~\ref{tab:coco}.

\begin{table}[ht]
\centering
\begin{tabular}{|l c c|}
\hline
\textbf{Model} & \textbf{mAP} & \textbf{mAP50} \\
\hline
\multicolumn{3}{c}{\textbf{EfficientNet}} \\
\hline
B-0 & 0.281 & 0.471 \\
B-1  & 0.242 & 0.403 \\
B-2  & 0.178 & 0.317 \\
B-3  & 0.124 & 0.237 \\
B-4  & 0.082 & 0.169 \\
B-5  & 0.077 & 0.161 \\
B-6  & 0.068 & 0.144 \\
B-7  & 0.059 & 0.127 \\
\hline
\multicolumn{3}{c}{\textbf{MobileNetV3}} \\
\hline
S  & 0.209
& 0.365	
\\
S-1  & 0.16  & 0.29 \\
S-2  & 0.132  & 0.246 \\
S-3  & 0.105  & 0.203 \\
S-4  & 0.088  & 0.177 \\
S-5  & 0.073  & 0.15 \\
S-6  & 0.055  & 0.119 \\
\hline
\end{tabular}
\caption{\textbf{COCO Detection mAP and mAP50 for EfficientNet and MobileNetV3 model families at input resolution 480$\times$384 except for EfficientNet and MobileNetV3}.}
\label{tab:coco}
\end{table}

\section{Detailed results}

\subsection{In-domain and out-domain results}
\textbf{In modality and RGB to IR cross-modality results:} Here we report tabular results for LLVIP, FLIR datasets. LLVIP results are presented in tables \ref{tab:llvip_effnet} and \ref{tab:llvip_mbnet} for EfficientNet and MobileNetV3 model families respectively. FLIR dataset's numbers are presented in tables \ref{tab:flir:EfficientNet} and \ref{tab:flir:Mobilenet}.
\textbf{Cross-dataset robustness results:} Here, we provide details of
results on FLIR$\rightarrow$LLVIP and LLVIP$\rightarrow$FLIR in RGB domain in \ref{tab:flir_llvip:EfficientNet},\ref{tab:flir_llvip:MobileNet},\ref{tab:llvip_flir:EfficientNet}, and \ref{tab:llvip_flir:MobileNet} respectively.
\textbf{DomainNet benchmark results:} Here, we provide details of
results obtained using EfficientNet and MobileNetV3 models in tables~\ref{tab:domainnet:effnet} and \ref{tab:domainnet:mbnet} respectively.
\textbf{ID-viewpoint, OOD-viewpoint In-domain and out-domain results}: Here, we provide details of performance of the models on Distech dataset are reported in tables \ref{tab:Distech:EfficientNet} and \ref{tab:Distech:Mobilenet} for EfficientNet and MobileNetV3 models respectively.
\end{document}